\documentclass{article} 
\usepackage{supporting_files/iclr2027_conference,times}

\usepackage{amsmath,amsfonts,bm}

\def\eqref#1{equation~\ref{#1}}

\def\1{\bm{1}}

\DeclareMathAlphabet{\mathsfit}{\encodingdefault}{\sfdefault}{m}{sl}
\SetMathAlphabet{\mathsfit}{bold}{\encodingdefault}{\sfdefault}{bx}{n}

\usepackage{amsmath}
\usepackage{amssymb}
\usepackage{amsthm}
\usepackage{graphicx}
\usepackage{booktabs}
\usepackage{array}
\usepackage{tabularx}
\usepackage{fancyvrb}
\usepackage{float}
\usepackage{xcolor}
\usepackage{multirow}
\DeclareUnicodeCharacter{2264}{\ensuremath{\leq}}
\DeclareUnicodeCharacter{2265}{\ensuremath{\geq}}
\usepackage{hyperref}
\usepackage{url}
\usepackage{titletoc}



\newtheorem{proposition}{Proposition}
\theoremstyle{definition}

\newif\ifshowflags
\showflagsfalse

\title{Know When to Abstain: Conformal Confidence Estimation for Text-to-SQL}

\author{Avinash Kumar Singh \\
Domyn \\
Hyderabad, India \\
\texttt{avinash.kumarsingh@domyn.com} \And
Bhaskarjit Sarmah \\
Domyn \\
New Delhi, India \\
\texttt{bhaskarjit.sarmah@domyn.com} \And
Stefano Pasquali \\
Domyn \\
New York, USA \\
\texttt{stefano.pasquali@domyn.com}}

\iclrfinalcopy 
\begin{document}

\maketitle

\begin{abstract}
Text-to-SQL systems often return queries that execute successfully but
answer the wrong question. Detecting these errors at deployment is hard
because execution correctness is defined against a gold query, which
exists only in labeled data. We formulate confidence estimation for
Text-to-SQL as probabilistic query validation under a latent
correctness variable that is observable at calibration time through
gold-SQL execution and latent at deployment. Our estimator separates
generation from validation. A generator LLM writes and executes the
query, and a \emph{different} critic LLM judges the query and its
execution result. The critic's validation and execution-grounded
alignment scores are fused with the generator's verbalized confidence,
token-level uncertainty, and SQL-structural features in a Random
Forest, using two sequential LLM calls. Two calibration steps give
finite-sample guarantees. A Learn-then-Test threshold ensures, with
probability $1-\delta$, that the error rate among accepted queries
stays below a user-chosen budget. A conformal p-value bounds the false
abstention rate. On Spider we use GPT-OSS-120B as generator and
Gemma-4-31B as critic. The estimator obtains an AUARC of 0.899.
Without abstention, 22.6\% of queries are wrong. With a 15\% risk
budget ($\delta{=}0.1$), a threshold is certified on all 10 resampled
splits, and the system accepts 52.4\% of queries on average with 9.1\%
observed error. Five different critic models give similar results. In a controlled comparison on
Spider with Olmo-3-32B as generator, the estimator obtains an AUARC of
89.5, compared with 81.2 for the strongest logit-based baseline. Applied to BIRD without retraining, the
confidence scores remain informative (AUROC 0.72, against 0.79 on
Spider), but the risk guarantee does not transfer, as expected without
exchangeability.
\end{abstract}

\section{Introduction}
\label{sec:intro}

LLMs reach high execution accuracy on Text-to-SQL
benchmarks~\citep{yu2018spider, li2023bird}, yet many of their errors
are queries that execute and return plausible but wrong
results~\citep{gong2026sqlens, somov2025confidence}. Such silent
failures pass into downstream reports without any execution-time
signal, and enterprise workloads with large schemas and
domain-specific semantics make them harder to
spot~\citep{lei2024spider, chen2024beaver}. A deployed system therefore
needs a per-query confidence score that decides whether to return a
result or abstain.

The difficulty is that correctness is defined relative to a gold
query. A generated query is correct when its execution result matches
that of the gold SQL. On labeled benchmarks this can be checked
exactly, but at deployment no gold SQL exists, so correctness is
\emph{latent}. Execution success is observable, but it is not
correctness. Any confidence score must therefore be inferred from
signals available at generation and execution time, and any
abstention rule must be fit on labeled data and then applied where no
labels exist.

Existing confidence estimators for Text-to-SQL mostly rely on one type
of signal. Sampling-based methods treat agreement among $K$ sampled
outputs as a proxy for correctness~\citep{wang2023selfconsistency,
kuhn2023semantic}. They cost $K$ LLM calls, and a model that
systematically misreads a schema can produce consistent but wrong
queries~\citep{stengel2023calibrated, ramachandran2024text}. Logit-based
methods aggregate token probabilities, possibly by SQL
clause~\citep{entezari2025confidence, ramachandran2024text}. They need a
single pass, but token probabilities reflect next-token uncertainty and
can remain high for fluent queries with the wrong
intent~\citep{stengel2023calibrated, desai2020calibration}. These
methods are usually evaluated by discrimination and calibration, and
with few exceptions~\citep{chen2025reliable} the abstention threshold
is not tied to a stated statistical property.

We treat confidence estimation as probabilistic query validation and
separate the roles of generation and validation. A generator LLM
writes the query and reports its own confidence. A \emph{different}
critic LLM then validates the query and checks whether its execution
result answers the question. Using a separate critic means the
judgment is not made by the model whose reasoning produced the error.
We combine the critic's validation and execution-grounded alignment
scores with the generator's verbalized confidence, token-level
uncertainty, and SQL-structural features. These signals take two
sequential LLM calls, one per model, plus local computation. A Random
Forest fuses them and is trained where correctness is observable.
The acceptance threshold is then chosen on a labeled calibration set
with Learn then Test~\citep{angelopoulos2022learn}. The result
guarantees, with high probability, that the error rate among accepted
queries stays below a budget the user chooses. A conformal p-value
computed from the same scores also bounds the fraction of correct
queries on which the system abstains.

On Spider, with GPT-OSS-120B as generator and Gemma-4-31B as critic,
a 15\% risk budget is certified on every resampled split. At that
budget the system accepts 52.4\% of queries with 9.1\% observed error,
against 22.6\% error without abstention. In a
controlled comparison with a shared generator, the estimator obtains
an AUARC of 89.5, compared with 81.2 for SAC-Product and 82.1 for
five-sample self-consistency. Our contributions are:
\begin{itemize}
  \item \textbf{Formulation.} We formalise confidence estimation for
  Text-to-SQL under latent execution correctness, separating
  what is observable at calibration time from what is latent at deployment. We state
  which selective-prediction quantities are guaranteed and which are
  empirical (Section~\ref{sec:problem}).
  \item \textbf{Estimator.} We propose a generator--critic estimator. A
  separate critic model validates the generator's query and its
  execution result, and a Random Forest fuses the critic's judgments
  with generator-side signals. An ablation shows that the LLM-derived
  signals carry most of the information, so a variant without logit
  access retains most of the performance
  (Sections~\ref{sec:method},~\ref{sec:ablation}).
  \item \textbf{Calibrated abstention.} We select the acceptance
  threshold with Learn then Test, which controls the error rate among
  accepted queries with high probability. We also derive a conformal
  bound on false abstention, and we characterise the resulting
  coverage--risk trade-off (Sections~\ref{sec:selective_pred},~\ref{sec:main_results}).
  \item \textbf{Evaluation.} We compare five critic models with
  self-critique. We also report a controlled baseline comparison,
  results for four generator backbones, 10-seed resampling, zero-shot
  transfer to BIRD and NL2SQL-BUGs, and a variant without logit access
  (Section~\ref{sec:experiments}).
\end{itemize}

\section{Related Work}
\label{sec:related}

\textbf{Error detection and confidence for Text-to-SQL.}
Prior work estimates query-level confidence from model probabilities
and their rescaling~\citep{ramachandran2024text,
stengel2023calibrated}, from SQL-aware logit aggregation and
execution-based grounding~\citep{entezari2025confidence,
somov2025confidence}, or from sub-clause frequencies with multivariate
Platt scaling~\citep{liu2025calibrating}. SQLens combines database and
LLM signals to detect and correct clause-level
errors~\citep{gong2026sqlens}, SQLFixAgent repairs queries through
multi-agent consistency checks~\citep{cen2025sqlfixagent}, and
NL2SQL-BUGs provides a benchmark for semantic error
detection~\citep{liu2025nl2sql}. These methods produce scores or
corrections. We combine several signal families into one estimator
and attach an abstention rule with a stated calibration property.

\textbf{Uncertainty estimation for LLMs.}
White-box methods use token-level
probabilities~\citep{hendrycks2017baseline, malinin2021uncertainty}.
Black-box methods use sampling consistency or semantic
equivalence~\citep{wang2023selfconsistency, kuhn2023semantic,
farquhar2024detecting}, whose underlying assumption is examined
by~\citet{xiao2025consistency}, or elicit verbalized
confidence~\citep{kadavath2022language, xiong2023confidence,
tian2023just}. We use verbalized and logit signals as input features
rather than as standalone scores.

\textbf{Selective prediction and conformal methods.}
Selective classification abstains on low-confidence inputs to lower
the error rate among accepted predictions~\citep{geifman2017selective};
\citet{wen2025know} survey abstention in LLMs. TrustSQL evaluates
Text-to-SQL reliability with penalty-based scoring that rewards
abstention~\citep{lee2024trustsql}, and \citet{chen2025reliable} use
conformal prediction to abstain during schema linking. Conformal
prediction gives finite-sample guarantees under
exchangeability~\citep{vovk2005algorithmic, angelopoulos2023conformal},
and has been applied to language generation~\citep{quach2023conformal}
and to LLMs without logit access~\citep{su2024api}. We control the
selective error rate with Learn then Test~\citep{angelopoulos2022learn},
which casts threshold selection as multiple hypothesis testing. Our
conformal confidence is the outlier-detection
construction~\citep{bates2023testing} with correct queries as the
reference class, and it controls false abstention.

\section{Problem Formulation}
\label{sec:problem}

\textbf{Setting.}
Let $x = (q, s)$ pair a natural-language question $q$ with a database
schema $s$, and let $D$ be the database instance. Throughout, $s$
denotes the schema, $\hat{s}$ the generated SQL query, and $s^*$ the
gold SQL query. Two models are involved. A \emph{generator}
$\theta_G$ defines an autoregressive distribution $P_{\theta_G}(\hat{s}
\mid x) = \prod_{t=1}^{T} P_{\theta_G}(\hat{s}_t \mid \hat{s}_{<t},
x)$, where $T$ is the number of tokens in $\hat{s}$. A decoding
procedure returns a single query $\hat{s}$, whose tokens $\tau =
(t_1,\dots,t_T)$ have probabilities $\pi_1,\dots,\pi_T$. A
\emph{critic} $\theta_C$, a different LLM from $\theta_G$, judges
$\hat{s}$ but does not generate it.

\textbf{Execution correctness as a latent variable.}
Let $\mathrm{Exec}(\hat{s}, D)$ denote deterministic execution. The
execution-correctness variable is
\begin{equation}
  Z(x, \hat{s}) = \mathbb{1}\!\left[\mathrm{Exec}(\hat{s}, D) = \mathrm{Exec}(s^*, D)\right].
  \label{eq:latent_z}
\end{equation}
The result $r = \mathrm{Exec}(\hat{s}, D)$ is always obtainable for
read-only \textsc{select} queries. $Z$ is latent because $s^*$ is
unavailable at deployment, not because execution is infeasible. On
labeled data, $Z$ is computed exactly by execution matching, which is
how we obtain training and calibration labels.

\textbf{Confidence function.}
A confidence function $C : \mathcal{X} \times \mathcal{Y} \to [0,1]$
maps $(x, \hat{s})$ to a score using only deployment-time signals: $q$,
$s$, $\hat{s}$, the token probabilities $\{\pi_t\}$, the execution
result $r$, the generator's verbalized confidence, and the critic's
judgments. Since $Z$ is unobserved at
deployment, we assess $C$ on labeled test data by discrimination
(ranking correct above incorrect queries), calibration (agreement of
scores with empirical correctness), and selective prediction.

\subsection{Selective Prediction and What Is Guaranteed}
\label{sec:selective_pred}

Given a confidence score $S(x,\hat{s})$ and a threshold $\lambda$, the
selective predictor is
\begin{equation}
  \mathcal{A}_\lambda(x) = \hat{s} \ \text{ if } S(x,\hat{s}) \geq \lambda,
  \qquad \mathcal{A}_\lambda(x) = \bot \ \text{(abstain) otherwise.}
  \label{eq:abstain}
\end{equation}
We use four population quantities:
\begin{align}
  \mathrm{Coverage}(\lambda) &= \mathbb{P}\!\left(S \geq \lambda\right),
  &
  \mathrm{Risk}(\lambda) &= \mathbb{P}\!\left(Z = 0 \mid S \geq \lambda\right),
  \label{eq:risk}\\
  \mathrm{FAR}(\lambda) &= \mathbb{P}\!\left(S < \lambda \mid Z = 1\right),
  &
  \mathrm{TAR}(\lambda) &= \mathbb{P}\!\left(S < \lambda \mid Z = 0\right).
  \label{eq:far}
\end{align}
FAR (false abstention rate) is the fraction of correct queries
withheld, and TAR (true abstention rate) the fraction of incorrect
queries withheld. With base accuracy $\rho = \mathbb{P}(Z{=}1)$, they
determine the other two quantities:
\begin{equation}
  \mathrm{Coverage}(\lambda) = \rho\,(1-\mathrm{FAR}) + (1-\rho)(1-\mathrm{TAR}),
  \qquad
  \mathrm{Risk}(\lambda) = \frac{(1-\rho)(1-\mathrm{TAR})}{\mathrm{Coverage}(\lambda)}.
  \label{eq:identity}
\end{equation}
Both calibration rules below use the fused estimate $p(x,\hat{s}) =
\hat{\mathbb{P}}(Z{=}1 \mid \mathbf{x})$ from Section~\ref{sec:stage4},
trained on a separate split, and a held-out calibration set
$\mathcal{C}$ of size $m$ with observed $Z_i$. Queries that fail to
execute are always rejected.

\textbf{Conformal confidence and false abstention.}
Let $\mathcal{C}_1 = \{i \in \mathcal{C} : Z_i = 1\}$ with $m_1 =
|\mathcal{C}_1|$, and define the nonconformity score $\eta = -\log p$.
The calibrated confidence is the conformal p-value
\begin{equation}
  \hat{c}(x,\hat{s})
  = \frac{1 + \bigl|\{ i \in \mathcal{C}_1 : \eta_i \geq \eta(x,\hat{s}) \}\bigr|}{m_1 + 1},
  \label{eq:conformal}
\end{equation}
with $\hat{c} = 0$ when $\hat{s}$ fails to execute. The \emph{conformal
rule} uses $S = \hat{c}$ and $\lambda = 1 - \alpha$ for an operating
level $\alpha \in (0,1]$.

\begin{proposition}[Finite-sample bound on false abstention]
\label{prop:conformal}
Suppose a correct test query $(x,\hat{s})$ with $Z=1$ is exchangeable
with $\{(x_i,\hat{s}_i)\}_{i \in \mathcal{C}_1}$, conditional on the
trained estimator $p$. Then $\mathbb{P}(\hat{c} \leq t \mid Z = 1) \leq t$
for every $t \in [0,1]$. Consequently, the conformal rule satisfies
\begin{equation}
  \mathrm{FAR}(1-\alpha) \;\leq\; 1-\alpha,
  \qquad\text{equivalently}\qquad
  \mathbb{P}\bigl(\hat{c} \geq 1-\alpha \mid Z = 1\bigr) \;\geq\; \alpha,
  \label{eq:far_guarantee}
\end{equation}
where the probability is over the calibration set and the test query.
\end{proposition}

This is the validity of conformal p-values used for outlier
detection~\citep[Sec.~4.4; see also \citealp{bates2023testing}]{angelopoulos2023conformal},
with correct queries as the reference class
(proof in Appendix~\ref{appendix:proofs}). By Eq.~\ref{eq:identity},
the conformal rule accepts about a fraction $\alpha$ of correct
queries, so a small $\alpha$ gives high precision at low coverage by
construction. It does not bound Risk.

\textbf{Risk control.}
To bound the error rate among accepted queries, the user fixes a risk
budget $\varepsilon$ and a failure probability $\delta$, and we select
the threshold with Learn then Test~\citep{angelopoulos2022learn}. Let
$\Lambda$ be a grid of thresholds on $p$ fixed before calibration. For
each $\lambda \in \Lambda$, let $n(\lambda)$ be the number of
calibration queries with $p_i \geq \lambda$, and $k(\lambda)$ the
number of those with $Z_i = 0$. We test $H_\lambda: \mathrm{Risk}(\lambda)
> \varepsilon$ with the binomial p-value $u_\lambda =
\mathbb{P}(\mathrm{Bin}(n(\lambda), \varepsilon) \leq k(\lambda))$, and
set $\hat{\lambda} = \min\{\lambda \in \Lambda : u_\lambda <
\delta/|\Lambda|\}$. The \emph{risk-controlled rule} uses $S = p$ and
$\lambda = \hat{\lambda}$. If no $\lambda$ qualifies, it abstains on
every query.

\begin{proposition}[Finite-sample risk control]
\label{prop:risk}
If the calibration and test queries are i.i.d.\ and $p$ is trained on
data disjoint from $\mathcal{C}$, then
\begin{equation}
  \mathbb{P}_{\mathcal{C}}\bigl(\mathrm{Risk}(\hat{\lambda}) \leq \varepsilon\bigr) \;\geq\; 1-\delta .
  \label{eq:risk_guarantee}
\end{equation}
\end{proposition}

Conditional on $n(\lambda)$, $k(\lambda)$ is binomial with success
probability $\mathrm{Risk}(\lambda)$, so $u_\lambda$ is a valid p-value
for $H_\lambda$~\citep[Sec.~5.5]{angelopoulos2023conformal}. The
Bonferroni correction controls the probability of certifying any
threshold whose risk exceeds $\varepsilon$
(Appendix~\ref{appendix:proofs}). Both propositions need gold SQL only
on the calibration set, and hold at deployment only while deployment
queries follow the calibration distribution (exchangeable for
Proposition~\ref{prop:conformal}, i.i.d.\ for Proposition~\ref{prop:risk}). Coverage under
the risk-controlled rule depends on how well $p$ separates correct
from incorrect queries. A tight budget may not be certifiable with few
calibration queries, in which case the rule abstains on every query.
\section{Method}
\label{sec:method}

\begin{figure*}[t]
\centering
\includegraphics[width=0.96\textwidth]{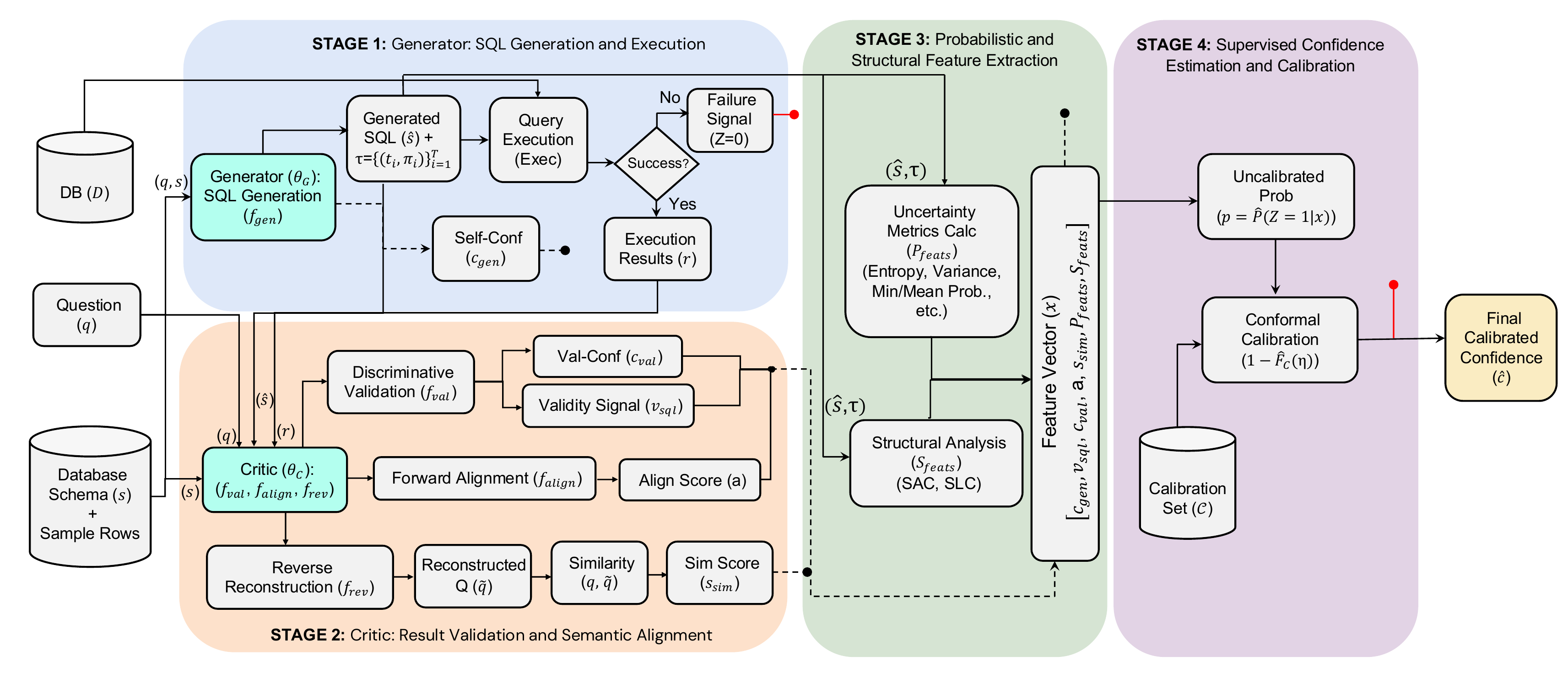}
\caption{The four-stage generator--critic estimator. Stage~1
(generator $\theta_G$, LLM call~1) generates $\hat{s}$ with its tokens
and probabilities $\tau$ and a self-confidence $c_{\mathrm{gen}}$,
then executes $\hat{s}$ on $D$. A failed execution sets $Z{=}0$ and
$\hat{c}{=}0$. Stage~2 (critic $\theta_C \neq \theta_G$, LLM call~2)
receives $(q, s, \hat{s}, r)$. It returns a validity verdict
$v_{\mathrm{sql}}$ with confidence $c_{\mathrm{val}}$, an alignment
score $a$, and the similarity $s_{\mathrm{sim}}$ between $q$ and the
question $\tilde{q}$ reconstructed from $r$. Stage~3 (local) computes
uncertainty and structural features from the generator's $(\hat{s},
\tau)$ and runs in parallel with Stage~2. Stage~4 (local) fuses
the feature vector $\mathbf{x}$ with a Random Forest and maps its
posterior to the calibrated confidence $\hat{c}$ using a calibration set
$\mathcal{C}$ labeled offline.}
\label{fig:pipeline}
\end{figure*}

The estimator uses two different LLMs with separate roles
(Figure~\ref{fig:pipeline}). The generator $\theta_G$ writes the query,
and the critic $\theta_C \neq \theta_G$ judges it. Stage~1 (generator,
LLM call~1) generates $\hat{s}$ with a verbalized confidence and token
probabilities, and executes $\hat{s}$. Stage~2 (critic, LLM call~2)
validates $\hat{s}$ and scores its execution result. Stage~3 (local)
computes probabilistic and structural features from the generator's
token probabilities. Stage~4 (local) fuses all features, applies
conformal calibration, and selects the risk-controlled threshold. Stages~2 and~3 run in parallel, so the pipeline
makes two sequential LLM calls, one to each model.
Table~\ref{tab:features} lists the feature families and the model or
stage that produces each.

\subsection{Stage 1: Generation and Execution (Generator)}
\label{sec:stage1}

The generator receives $(q, s)$, where the schema includes sample
rows, and a generation prompt (Appendix~\ref{stage_1_prompt}). Before
answering, it checks its query internally for question alignment,
grounding in the schema, literal values, SQLite syntax, logical
correctness, result sanity, and minimality. Its output contains only
$\hat{s} = f_{\mathrm{gen}}(q,s)$ and a verbalized confidence
$c_{\mathrm{gen}} \in [0,100]$. The prompt asks for a confidence that
reflects the weakest critical aspect of the query, not its syntactic
validity. The generator returns no validation verdict. Verdicts come
only from the critic. We record the probabilities $\pi_t$ of the
tokens of $\hat{s}$ for Stage~3. We then execute $\hat{s}$ on $D$ to
obtain $r = \mathrm{Exec}(\hat{s}, D)$. If execution fails, $Z{=}0$ is
certain, and the query receives $\hat{c}{=}0$ without further
processing.

\subsection{Stage 2: Critic Validation and Execution-Grounded Alignment (Critic)}
\label{sec:stage2}

The critic receives $(q, s, \hat{s}, r)$ in a single prompt
(Appendix~\ref{stage_2_prompt}). It does not see $c_{\mathrm{gen}}$ or
the generator's reasoning, and it has no access to gold SQL or gold
answers. It judges the executed result against the question, not
merely whether the values look plausible. Verbalized judgments can
surface semantic problems that token probabilities do
not~\citep{kadavath2022language, xiong2023confidence}.

\textbf{Discriminative validation.} The critic checks whether $r$
correctly and completely answers $q$. The checks cover required
columns and rows, the requested operation or aggregation, filters and
comparison operators, ordering and top-$k$ limits, grouping
granularity, date and time constraints, the consistency of returned
values, and completeness. Violating any substantive requirement makes
the result incorrect. The critic returns a binary verdict
$v_{\mathrm{sql}} = f_{\mathrm{val}}(q, s, \hat{s}, r) \in \{0,1\}$ and
its confidence in that verdict, $c_{\mathrm{val}} \in [0,100]$.

\textbf{Execution-grounded alignment.} Ignoring $q$, the critic
reconstructs the most precise question $\tilde{q} = f_{\mathrm{rev}}(s, r)$
that $r$ answers, using only $r$ and the schema, without repairing
missing conditions. It also returns an alignment score
$a = f_{\mathrm{align}}(q, s, \hat{s}, r) \in [0,100]$, which measures
how well the result answers $q$ in terms of the
requested information, filters, aggregation, ordering, grouping, dates,
and completeness. The critic also returns the semantic similarity
$s_{\mathrm{sim}} = \mathrm{Sim}(q,\tilde{q}) \in [0,100]$. The prompt
caps $a$ and $s_{\mathrm{sim}}$ when $v_{\mathrm{sql}} = 0$. A low
$a$ or $s_{\mathrm{sim}}$ flags results that are topically related but
answer a different question.

\subsection{Stage 3: Probabilistic and Structural Features}
\label{sec:stage3}

From the generator's token probabilities $\{\pi_t\}_{t=1}^T$ of
$\hat{s}$, we compute without further LLM calls:
\begin{align}
  \mathrm{FTC} &= \textstyle\prod_{t} \pi_t, \quad
  \pi_{\min} = \min_t \pi_t, \quad
  \mu_{\mathrm{arith}} = \tfrac{1}{T}\textstyle\sum_t \pi_t, \quad
  \mu_{\mathrm{geom}} = \bigl(\textstyle\prod_t \pi_t\bigr)^{1/T},
  \label{eq:ftc}\\
  H &= -\textstyle\sum_t \pi_t \log \pi_t, \quad
  \sigma^2 = \tfrac{1}{T}\textstyle\sum_t(\pi_t-\mu_{\mathrm{arith}})^2, \quad
  \ell_{\mathrm{CLM}} = -\textstyle\sum_t \log \pi_t .
  \label{eq:entropy}
\end{align}
We also recompute the first five statistics on SQL keyword and
non-keyword tokens separately, so that uncertainty confined to
identifiers or literals is not averaged away. We add the probability
range and a 13-bin probability histogram. Token probabilities do not distinguish schema
identifiers from other tokens, so we add SQL-aware
features~\citep{entezari2025confidence, somov2025confidence}. Let
$\mathcal{G}$ partition the tokens of $\hat{s}$ into clause groups
(\textsc{select}, \textsc{where}, table identifiers, and so on), and
let $\tau_{\mathrm{sch}}$ be the tokens that name schema
elements. We compute
\begin{equation}
  \mathrm{SAC}_{\mathrm{geom}}
  = \Bigl(\textstyle\prod_{g \in \mathcal{G}}
    \tfrac{1}{|g|}\sum_{t \in g} \pi_t\Bigr)^{1/|\mathcal{G}|},
  \qquad
  \mathrm{SLC}
  = \bigl(\textstyle\prod_{t \in \tau_{\mathrm{sch}}} \pi_t\bigr)^{1/|\tau_{\mathrm{sch}}|}
  \ \ (\mathrm{SLC}{=}0 \text{ if } \tau_{\mathrm{sch}}{=}\emptyset).
  \label{eq:sac}
\end{equation}
SLC targets schema grounding, a primary failure mode in cross-domain
Text-to-SQL~\citep{yu2018spider, li2023bird}. A schema-table mean and
a clause-consistency score (C3S) complete this family. The full vector
has 41 features (Table~\ref{tab:features}; each one is listed in
Appendix~\ref{appendix:feature_list}). Some are near-duplicates, for
example $\ell_{\mathrm{CLM}} = -\log \mathrm{FTC}$, and a tree ensemble
is unaffected by such redundancy.

\begin{table}[t]
\centering
\small
\caption{Features in $\mathbf{x}$, the model and stage that produce
them, and the failure mode they target. $\theta_G$: generator;
$\theta_C$: critic. The last row is listed in full in
Appendix Table~\ref{tab:feature_list}.}
\label{tab:features}
\setlength{\tabcolsep}{4pt}
\resizebox{\linewidth}{!}{%
\begin{tabular}{lllcl}
\toprule
\textbf{Family} & \textbf{Features} & \textbf{Source} & \textbf{Dim} &
\textbf{Targeted failure mode} \\
\midrule
Verbalized   & $c_{\mathrm{gen}}$
             & Stage 1, $\theta_G$ & 1  & Generator uncertainty \\
             & $v_{\mathrm{sql}},\, c_{\mathrm{val}}$
             & Stage 2, $\theta_C$ & 2  & Result violates question \\
Alignment    & $a,\, s_{\mathrm{sim}}$
             & Stage 2, $\theta_C$ & 2  & Plausible but wrong results \\
Logit ($P_{\mathrm{feats}}$) & $\mathrm{FTC}, \pi_{\min}, \mu_{\mathrm{arith}},
               \mu_{\mathrm{geom}}, H, \sigma^2, \ell_{\mathrm{CLM}}$
             & Stage 3, local & 7  & Token-level uncertainty \\
SQL-structural ($S_{\mathrm{feats}}$) & $\mathrm{SAC}_{\mathrm{geom}},\, \mathrm{SLC}$
             & Stage 3, local & 2  & Schema grounding errors \\
Variants     & Subset statistics, range, histogram,
             & Stage 3, local & 27 & Localised uncertainty \\
             & \quad schema-table mean, C3S & & & \\
\midrule
\textbf{Total} & & & \textbf{41} & \\
\bottomrule
\end{tabular}}
\end{table}

\subsection{Stage 4: Supervised Fusion and Conformal Calibration}
\label{sec:stage4}

A Random Forest $f_{\mathrm{RF}}$ maps $\mathbf{x}$ to $p =
\hat{\mathbb{P}}(Z{=}1 \mid \mathbf{x})$. It is trained on labeled
queries whose $Z$ is computed by execution matching. The forest
handles features on different scales without normalisation and can
use feature interactions, for example high FTC with low alignment. No
single feature separates the two classes well
(Appendix~\ref{appendix:feature_distribution}). On a disjoint
calibration split, we compute $\eta_i = -\log p_i$ for correct queries
and use Eq.~\ref{eq:conformal} to map $p$ to $\hat{c}$. Because
$\hat{c}$ is a non-decreasing function of $p$, calibration preserves the
ranking of queries up to ties, and we report AUROC and AUARC of $p$. The three
phases are separated as follows. Training and calibration use gold SQL
to compute $Z$. Deployment uses only the generator, the critic, the
local Stage~3 features, the frozen forest, and the stored calibration
scores.
\section{Experiments}
\label{sec:experiments}

\subsection{Experimental Setup}
\label{sec:setup}

\textbf{Data and models.}
We use 3,181 question--SQL pairs from the Spider dev and test
sets~\citep{yu2018spider}, covering 200 databases. After removing pairs
whose gold SQL fails to execute, 3,175--3,179 remain, depending on
the generation run. For zero-shot transfer
we use the BIRD dev set~\citep{li2023bird} (1,534 samples) and
NL2SQL-BUGs~\citep{liu2025nl2sql} (1,019 correct and 999 semantically
erroneous queries). Model details are in
Appendix~\ref{appendix:model_details}.

\textbf{Generator and critic models.} Different models always serve in
Stages~1 and~2, and Gemma-4-31B-it is the critic unless stated
otherwise. The main configuration uses GPT-OSS-120B as generator. The
four-backbone study uses GPT-OSS-120B, Olmo-3-32B-Think,
Qwen2.5-72B-Instruct, and Llama-3.3-Nemotron-Super-49B-v1.5 as
generators. The controlled comparison and the ablations use
Olmo-3-32B-Think. To test the choice of critic, we also run
Qwen3.8-27B, DeepSeek-V4.1-Flash, Nemotron-3-Super, and
Qwen2.5-72B-Instruct as critics for GPT-OSS-120B. As a reference, we
report self-critique, where the generator acts as its own critic
($\theta_C = \theta_G$).

\textbf{Protocol.}
\label{data_splits}
Spider is split, stratified by $Z$, into training (60\%), calibration
(20\%), and test (20\%) sets, for example 1,907 / 636 / 636 queries. For each
configuration, a separate Random Forest is trained on features computed
from that configuration's outputs, with hyperparameters chosen by 5-fold
cross-validated random search on the training split
(Appendix~\ref{appendix:random_forest}). Conformal calibration uses the
calibration split, and all metrics are computed on the test split. For
zero-shot transfer, forests and calibration scores fit on Spider are
applied unchanged to BIRD and NL2SQL-BUGs (Section~\ref{sec:transfer}). No target-domain data are used for training or
calibration.

\textbf{Metrics.}
\emph{Discrimination}: AUROC, and AUARC (area under the
accuracy--rejection curve, our primary metric). \emph{Calibration}: ECE
(10 equal-width bins), ACE (equal-mass bins), and Brier score.
\emph{Selective prediction}: Coverage, Risk, TAR, and FAR
(Eqs.~\ref{eq:risk}--\ref{eq:far}), estimated on the test split, and
conditional accuracy $\mathrm{CondAcc} = 1 - \mathrm{Risk}$, the
accuracy among accepted queries. Risk is covered by
Proposition~\ref{prop:risk} for the risk-controlled rule. FAR is
covered by Proposition~\ref{prop:conformal} for the conformal rule.
Unless stated otherwise, we use $\delta = 0.1$ and a grid $\Lambda =
\{0.01, 0.02, \dots, 0.99\}$.
Appendix~\ref{eval_metric} gives full definitions. Per-query latency is
reported in Appendix~\ref{latency_breakdown}.

\subsection{Generator--Critic Results}
\label{sec:main_results}

\begin{table}[t]
\centering
\small
\caption{Critic comparison on the Spider test split ($n{=}636$,
generator GPT-OSS-120B). Selective metrics use the conformal rule at
$\alpha{=}0.1$, for which Proposition~\ref{prop:conformal} bounds FAR
$\leq 0.9$ in expectation. The self-critique run uses a separate generation pass, so its
base accuracy differs.}
\label{tab:critic}
\resizebox{\linewidth}{!}{%
\begin{tabular}{lcccccc}
\toprule
 & Self-critique & \multicolumn{5}{c}{Critic} \\
\cmidrule(lr){3-7}
 & ($\theta_C = \theta_G$) & Gemma-4-31B & Qwen3.8-27B & DeepSeek-V4.1 & Nemotron-3-S & Qwen2.5-72B \\
\midrule
Base accuracy $\rho$ & 0.786 & 0.774 & 0.774 & 0.774 & 0.774 & 0.774 \\
ECE $\downarrow$ & 0.149 & 0.158 & 0.155 & 0.156 & 0.142 & 0.166 \\
AUROC $\uparrow$ & 0.781 & 0.785 & 0.840 & 0.811 & 0.779 & 0.826 \\
AUARC $\uparrow$ & 0.913 & 0.899 & 0.922 & 0.916 & 0.901 & 0.915 \\
\midrule
Coverage & 0.077 & 0.079 & 0.063 & 0.080 & 0.074 & 0.082 \\
CondAcc $=1-$Risk & 0.980 & 0.940 & 1.000 & 1.000 & 0.957 & 0.981 \\
TAR & 0.993 & 0.979 & 1.000 & 1.000 & 0.986 & 0.993 \\
FAR (bound: $\leq 0.9$) & 0.904 & 0.904 & 0.919 & 0.896 & 0.909 & 0.896 \\
\bottomrule
\end{tabular}}
\end{table}

Table~\ref{tab:critic} compares five critics with self-critique.
AUARC ranges from 0.899 (Gemma-4-31B) to 0.922 (Qwen3.8-27B), and
self-critique obtains 0.913. These differences are of the same order
as the 10-seed variability (Section~\ref{sec:robustness}), so a
separate critic keeps discrimination while removing self-judgment.
Under the conformal rule at $\alpha{=}0.1$, every configuration accepts
6.3--8.2\% of queries, 94.0--100\% of them correct, with FAR
(0.896--0.919) close to its bound of 0.9. Partial configurations are in
Appendix~\ref{appendix:partial_critic}. Across four generator
backbones with the Gemma critic, AUARC ranges from 0.895 to 0.918
(Appendix Table~\ref{tab:full_multillm}).

\textbf{Risk-controlled abstention.} The conformal rule at
$\alpha{=}0.1$ accepts few queries by construction, since it fixes the
accepted fraction of correct queries at about $\alpha$
(Section~\ref{sec:selective_pred}). The risk-controlled rule instead
takes the most lenient threshold for which the calibration data
certify $\mathrm{Risk} \leq \varepsilon$. Table~\ref{tab:ltt} reports
it over 10 resampled splits. At $\varepsilon{=}0.15$, a threshold is
certified on all 80 splits of the eight configurations, and the system
accepts 38.8--64.1\% of queries on average. With the Gemma critic it
accepts 52.4\%, compared with 7.9\% under the conformal rule. At
$\varepsilon{=}0.20$, coverage reaches 67.5--87.7\%. Across all 194
certified runs, the test-split risk never exceeds $\varepsilon$. A
budget of $\varepsilon{=}0.10$ is certified on only 0--8 of 10 splits. With 636
calibration queries, the binomial bound is too wide for the small
accepted sets at the strict end. The low coverage of the conformal
rule therefore reflects the rule, not the estimator
(sweep in Appendix~\ref{appendix:operating_points}).

\begin{table}[t]
\centering
\small
\caption{Risk-controlled abstention (Proposition~\ref{prop:risk},
$\delta{=}0.1$) on Spider over 10 resampled splits. Cert.: splits on
which a threshold is certified; otherwise the system abstains on every
query. Cov.: coverage (\%) averaged over all 10 splits, counting
uncertified splits as zero. Risk: test-split error among accepted
queries (\%), mean (max) over certified splits. Critic rows use
generator GPT-OSS-120B; generator rows use the Gemma critic.
$^\ast$Generation run of the four-backbone study.}
\label{tab:ltt}
\setlength{\tabcolsep}{2.8pt}
\begin{tabular}{l ccc ccc ccc}
\toprule
 & \multicolumn{3}{c}{$\varepsilon = 0.10$} & \multicolumn{3}{c}{$\varepsilon = 0.15$} & \multicolumn{3}{c}{$\varepsilon = 0.20$} \\
\cmidrule(lr){2-4}\cmidrule(lr){5-7}\cmidrule(lr){8-10}
 & Cert. & Cov. & Risk & Cert. & Cov. & Risk & Cert. & Cov. & Risk \\
\midrule
Critic: Gemma-4-31B          & 0/10 & 0.0 & --- & 10/10 & 52.4 & 9.1 (11.7) & 10/10 & 74.2 & 14.1 (16.1) \\
Critic: Qwen3.8-27B          & 7/10 & 27.6 & 4.7 (8.0) & 10/10 & 64.1 & 9.9 (14.6) & 10/10 & 79.8 & 14.2 (16.0) \\
Critic: DeepSeek-V4.1        & 6/10 & 21.9 & 4.9 (9.1) & 10/10 & 55.4 & 8.7 (11.9) & 10/10 & 75.3 & 14.0 (15.9) \\
Critic: Nemotron-3-Super     & 6/10 & 19.6 & 5.0 (8.2) & 10/10 & 54.2 & 9.3 (11.4) & 10/10 & 72.7 & 14.6 (17.2) \\
Gen.: Olmo-3-32B             & 1/10 & 1.9 & 7.4 (7.4) & 10/10 & 38.8 & 8.2 (12.1) & 10/10 & 67.5 & 14.1 (17.8) \\
Gen.: Qwen2.5-72B            & 4/10 & 14.4 & 5.9 (7.8) & 10/10 & 62.2 & 9.7 (12.6) & 10/10 & 87.7 & 15.5 (17.0) \\
Gen.: Llama-49B              & 2/10 & 7.2 & 7.0 (7.1) & 10/10 & 51.9 & 8.6 (11.5) & 10/10 & 74.1 & 14.3 (16.0) \\
Gen.: GPT-OSS-120B$^\ast$    & 8/10 & 23.4 & 4.1 (6.9) & 10/10 & 57.1 & 9.3 (12.7) & 10/10 & 76.6 & 14.2 (16.4) \\
\bottomrule
\end{tabular}
\end{table}

\subsection{Controlled Comparison Against Baselines}
\label{sec:controlled}

All methods in Table~\ref{tab:controlled} are re-implemented on
Olmo-3-32B-Think with the same splits and evaluation code. Baselines
are SAC-Product~\citep{entezari2025confidence} and Pooled
Token-Level~\citep{ramachandran2024text} (logit-based),
Self-Consistency with $K{=}3,5$~\citep{wang2023selfconsistency}, and
Single CoT Verify, which uses the same generator and critic calls
without feature fusion or conformal calibration. Numbers reported in the original papers use
other backbones and protocols. We list them in
Appendix~\ref{appendix:compare_with_existing_methods} for reference
only.

\begin{table}[t]
\centering
\small
\caption{Controlled comparison on Spider. All methods use
Olmo-3-32B-Think as generator and share splits and evaluation code.
$\dagger$: logit-based. Calls: LLM calls per query. Per-query latency
of our pipeline is reported in Appendix~\ref{latency_breakdown}.}
\label{tab:controlled}
\setlength{\tabcolsep}{4pt}
\begin{tabular}{lcccc}
\toprule
\textbf{Method} & \textbf{Calls} & \textbf{AUARC} & \textbf{AUROC}
  & \textbf{ECE} \\
\midrule
Self-Consistency ($K{=}3$) & 3 & 79.3 & 68.4 & 0.198 \\
Self-Consistency ($K{=}5$) & 5 & 82.1 & 71.2 & 0.183 \\
Single CoT Verify          & 2 & 78.9 & 63.6 & 0.030 \\
SAC-Product$\dagger$~\citep{entezari2025confidence}
                           & 1 & 81.2 & 72.4 & 0.161 \\
Pooled Token-Level$\dagger$~\citep{ramachandran2024text}
                           & 1 & 79.7 & 70.8 & 0.174 \\
\midrule
\textbf{Ours} & \textbf{2} & \textbf{89.5} & \textbf{81.6}
  & 0.126 \\
\bottomrule
\end{tabular}
\end{table}

Our estimator obtains an AUARC of 89.5, compared with 81.2 for
SAC-Product and 82.1 for Self-Consistency ($K{=}5$). A bootstrap over
the test split ($B{=}1{,}000$) gives a 95\% interval of $(86.7,
91.9)$, whose lower bound lies above every baseline in the table.
Single CoT Verify uses the same two calls and obtains 78.9. The
10.6-point gap therefore comes from feature fusion, not from
additional LLM calls. Self-Consistency with five calls remains 7.4
points lower, which is consistent with agreement among samples failing
to detect systematic errors~\citep{kuhn2023semantic,
ramachandran2024text}. Single CoT Verify has the lowest ECE but ranks
queries poorly; our method targets ranking and risk control.

\subsection{Ablation Study}
\label{sec:ablation}

\begin{table}[t]
\centering
\small
\begin{minipage}[t]{0.56\textwidth}
\centering
\caption{Leave-one-family-out ablation (generator
Olmo-3-32B-Think, critic Gemma-4-31B, Spider). 10 seeds: mean $\pm$
s.d.\ over resampled splits. ECE of the forest posterior $p$.}
\label{tab:ablation}
\setlength{\tabcolsep}{3pt}
\begin{tabular}{lccc}
\toprule
Configuration & AUARC & 10 seeds & ECE \\
\midrule
Full model                          & 89.5 & $88.5 \pm 0.8$ & 0.126 \\
w/o Verbalized                      & 87.6 & $87.6 \pm 0.8$ & 0.124 \\
w/o Alignment                       & 87.8 & $87.4 \pm 0.8$ & 0.134 \\
w/o Logit $P_{\mathrm{feats}}$      & 90.5 & $88.3 \pm 0.9$ & 0.143 \\
w/o Structural $S_{\mathrm{feats}}$ & 89.0 & $88.1 \pm 0.8$ & 0.122 \\
Logit-free (5 LLM scores)           & 88.6 & $86.9 \pm 1.2$ & 0.187 \\
\midrule
\textit{SAC-Product (baseline)}     & \textit{81.2} & --- & \textit{0.161} \\
\bottomrule
\end{tabular}
\end{minipage}\hfill
\begin{minipage}[t]{0.41\textwidth}
\centering
\caption{Incremental build-up from Single CoT Verify (same setting),
retraining the forest at each row. All additions are local.}
\label{tab:twocall}
\setlength{\tabcolsep}{3pt}
\resizebox{\linewidth}{!}{%
\begin{tabular}{lcc}
\toprule
Configuration & AUARC & $\Delta$ \\
\midrule
Single CoT Verify (2 calls)          & 78.9 & ---    \\
Forest on verbalized scores          & 84.5 & $+5.6$ \\
+ Logit/structural features          & 87.9 & $+9.0$ \\
+ Alignment (Full)                   & 89.5 & $+10.6$ \\
\bottomrule
\end{tabular}}
\end{minipage}
\end{table}

Table~\ref{tab:ablation} removes one feature family at a time and
retrains the forest. Removing the verbalized or alignment family
lowers AUARC by 0.9 and 1.1 points over 10 seeds. Removing the logit or
structural family changes AUARC by less than the seed variability, and
on the single split removing the logit family even raises it. With a
separate critic, the LLM-derived families carry most of the signal.
Conformal calibration does not change the ranking (a monotone map); it
provides the thresholds of Propositions~\ref{prop:conformal}
and~\ref{prop:risk}. Since $\hat{c}$ is a conformal p-value, not a
probability, we report ECE for $p$ only.

In Table~\ref{tab:twocall}, fusing the three verbalized scores raises
AUARC from 78.9 to 84.5, token-level and structural features add 3.4
points, and alignment adds 1.6.

\textbf{Logit-free variant.} Closed APIs often do not expose logits.
Keeping only the five LLM-derived scores (no token-based feature)
gives an AUARC of 88.6, 0.9 points below the full model (1.6 over 10
seeds) and 7.4 above SAC-Product, which itself needs logits. We
emulate this setting and have not evaluated closed-API models directly.

\subsection{Seed Variability and Signal Reliability}
\label{sec:robustness}

We resampled the splits 10 times with hyperparameters fixed, for all
eight configurations of Table~\ref{tab:ltt}
(Appendix~\ref{appendix:seeds}). The 10-seed mean AUARC lies within
0.013 of the single-split value in each case. For the Gemma critic,
AUARC is $0.900 \pm 0.010$. Single-split AUROC differs from the 10-seed mean
by up to 0.029, so AUROC differences of that size should not be over-interpreted. Mean FAR at
$\alpha{=}0.1$ is 0.889--0.907, close to the bound of
Proposition~\ref{prop:conformal}. Each LLM-derived score has a
standalone AUROC between 0.525 and 0.771, above the 0.5 chance level,
for all four generators and all four critics
(Appendix~\ref{appendix:score_reliability}), and below the forest's
AUROC of 0.765--0.840. Appendix~\ref{appendix:error_analysis} analyses
the errors that the critic misses.

\subsection{Zero-Shot Cross-Dataset Transfer}
\label{sec:transfer}

We apply the Spider-fit forests and calibration scores to BIRD and
NL2SQL-BUGs without retraining or recalibration. The target data are
not exchangeable with Spider, so neither proposition applies.

\textbf{BIRD.} BIRD has more complex joins, nested aggregations, and
longer schemas than Spider, and only 34.2\% of generated queries are
correct. The estimator obtains an AUROC of 0.724 (0.791 in-domain) and
an AUARC of 0.515, where random and perfect rankings give 0.343 and
0.709. Its lower ECE than on Spider (0.091 vs 0.133) reflects the base
rate, not better calibration (Appendix~\ref{appendix:bird_ece}). The
risk threshold certified on Spider does not carry over: at
$\varepsilon{=}0.15$ the observed BIRD risk is 30.1\%, which shows why
Proposition~\ref{prop:risk} requires i.i.d.\ data.

\textbf{NL2SQL-BUGs.} This benchmark supplies SQL written by other
systems, so only Stage~2 applies. \emph{Our Stage~2 critic alone, with an
open 31B model and execution grounding, reaches 77.4\% accuracy on
NL2SQL-BUGs, on par with the best models reported
by~\citet{liu2025nl2sql} (76.8\%)}, although prompts, models, and inputs
differ. A Spider-trained forest on the critic's scores transfers less
well (F1 0.708 vs.\ 0.779 for the verdict), as generator-side features
are missing and the error rate differs.

\section{Conclusion}
\label{sec:conclusion}

We studied confidence estimation for Text-to-SQL when execution
correctness is observable on labeled data but latent at deployment.
A generator writes and executes the query, a different critic model
validates it, and a Random Forest fuses both models' signals. A
Learn-then-Test threshold controls the error rate among accepted
queries with high probability, and a conformal p-value bounds false
abstention. With GPT-OSS-120B as generator and Gemma-4-31B as critic,
a 15\% risk budget is certified on every resampled split of Spider,
and the system accepts 52.4\% of queries at 9.1\% observed error. In a controlled comparison with a
shared generator, the estimator improves AUARC by 8.3 points over the
strongest logit-based baseline. Five critic models give similar
results, and a variant without logit access loses 1.6 AUARC. On BIRD, the Spider-trained estimator still separates correct from
incorrect queries without retraining (AUROC 0.72), but the Spider risk threshold no longer holds.

\textbf{Limitations.}
\textbf{(1)~Scope of the guarantees.} Both propositions assume that
deployment queries follow the calibration distribution, and neither
holds under distribution shift. Coverage is not guaranteed, and with
636 calibration queries a 10\% risk budget is often not certifiable.
\textbf{(2)~Dependence on LLM judgments.} The largest feature families
come from LLM judgments. A separate critic removes self-judgment, but
the critic is itself an LLM, and errors that it misreads remain hard
to detect (Appendix~\ref{appendix:error_analysis}).
\textbf{(3)~Logit access and overhead.} The full model needs the
generator's token probabilities and two sequential LLM calls, about 4.3\,s per query in our
setup (Appendix~\ref{latency_breakdown}).
\textbf{(4)~Evaluation scope.} We evaluate SQLite benchmarks and
open-weight models; Spider~2.0~\citep{lei2024spider} and other dialects
remain untested. Only the alignment step and the structural features
depend on SQL; code with unit tests has an execution analogue, but we
have not evaluated other tasks (Appendix~\ref{appendix:generalization}).

\subsection*{AI use statement}
We used Large Language Models (LLMs) to correct grammar and to assist with
writing. LLMs were not involved in any part of the research discussions, analyses, or idea generation. All insights, contributions, and intellectual content are entirely the authors’ own.The language models evaluated in this paper, as generators and
critics, are objects of study, not authoring tools. 

\subsection*{Ethics statement}

This work uses public benchmarks (Spider, BIRD, NL2SQL-BUGs) that
contain no personal data, and open-weight models. Its aim is to reduce
silently wrong database answers reaching users. A deployment that
treats accepted queries as verified could still act on the remaining
errors. The error rate among accepted queries is bounded only with
probability $1-\delta$ and under i.i.d.\ sampling
(Section~\ref{sec:selective_pred}).

\subsection*{Reproducibility statement}

Section~\ref{sec:problem} states the formal setting and assumptions.
Section~\ref{sec:setup} describes the data, splits, and protocol.
Appendix~\ref{stage_1_prompt} and \ref{stage_2_prompt} give the full prompts, Appendix~\ref{appendix:random_forest} the forest
hyperparameters, Appendix~\ref{eval_metric} the metric definitions,
and Appendix~\ref{appendix:model_details} the model checkpoints.
Code for feature extraction, training, calibration, and evaluation
will be released upon paper acceptance.

\bibliography{supporting_files/references}
\bibliographystyle{supporting_files/iclr2027_conference}

\clearpage
\appendix
\newcolumntype{L}[1]{>{\raggedright\arraybackslash}p{#1}}
\newcolumntype{Y}{>{\raggedright\arraybackslash}X}

\begingroup
\hypersetup{pdfborder={0 0 0}}
\definecolor{toclink}{RGB}{31,111,235}
\titlecontents{section}[1.8em]{\addvspace{1.1em}\bfseries}%
  {\color{toclink}\contentslabel{1.8em}}{\hspace*{-1.8em}\color{toclink}}%
  {\hfill\color{black}\contentspage}
\titlecontents{subsection}[4.3em]{\addvspace{0.25em}}%
  {\color{toclink}\contentslabel{2.5em}}{\hspace*{-2.5em}\color{toclink}}%
  {\color{black}\titlerule*[0.75pc]{.}\contentspage}
\begin{center}
{\Large\bfseries Appendix}
\end{center}
\vspace{0.5em}
\startcontents[appendix]
\printcontents[appendix]{}{1}{\setcounter{tocdepth}{2}}
\endgroup
\clearpage
\section{Prompts}

\subsection{Stage 1 Prompt (Generator)}
\label{stage_1_prompt}

The generator $\theta_G$ receives only the question and the schema with
sample rows. It validates its query internally but outputs only the
query and its confidence $c_{\mathrm{gen}}$ (\texttt{confidence}), with
no validation verdict. Long lines are wrapped for display.

\begin{Verbatim}[fontsize=\small]
You are a Text-to-SQL expert and a strict SQL correctness auditor.

Your task is to generate ONE SQLite-compatible SQL query that answers the
  given
natural-language question using ONLY the provided database schema and
  examples.

You must internally perform both SQL generation and rigorous correctness
  validation, but
your final output MUST contain only the SQL query and its confidence score.

--------------------
INPUTS
--------------------

Natural Language Question:
<QUESTION>

Database Schema with Examples:
<TABLE_SCHEMA>

--------------------
SQL GENERATION AND INTERNAL VALIDATION
--------------------

Generate a SQLite-compatible SQL query that correctly answers the question.

Before assigning confidence, internally verify ALL of the following:

1. QUESTION ALIGNMENT
- The SQL must directly answer the intent of the question.
- Ensure every selected column, aggregation, filter, grouping, ordering, join,
  and limit
  is justified by the question.
- Do not answer a different or narrower interpretation of the question.
- Do not introduce unnecessary joins, filters, aggregations, or conditions.

2. CONTEXT-ONLY GENERATION
- Use ONLY tables, columns, relationships, values, and concepts explicitly
  available in
  the provided schema and examples.
- Do NOT hallucinate tables, columns, relationships, categorical values,
  identifiers,
  dates, entities, or business concepts.
- Do NOT assume information that is not supported by the provided context.
- Do NOT invent literal filter values.
- If a required value is not supported by the question or schema/examples, do
  not
  fabricate it.

3. VALUE AND LITERAL VALIDATION
- Every literal value used in WHERE, HAVING, JOIN, CASE, IN, or other
  conditions must be
  supported by the question or the provided schema/examples.
- Preserve the exact meaning and representation of values supported by the
  context.
- Do not translate, normalize, or reinterpret categorical values unless the
  provided
  context supports doing so.
- If the question contains a value, use it faithfully unless the schema
  clearly requires
  a supported representation.
- Never introduce a value merely because it seems plausible.

4. SQL CORRECTNESS
- The query must follow valid SQLite syntax.
- Use SQLite-supported functions, operators, expressions, date handling,
  aggregation,
  joins, subqueries, CTEs, and window functions only.
- Do not use database-specific syntax from PostgreSQL, MySQL, SQL Server,
  Oracle, or
  other SQL dialects when it is not valid SQLite.
- Ensure table aliases, column references, GROUP BY, HAVING, ORDER BY, JOIN
  conditions,
  subqueries, and CTEs are syntactically and logically valid.
- Ensure referenced tables and columns actually exist in the provided schema.
- The SQL must be complete and executable.

5. LOGICAL CORRECTNESS
- Check that the SQL computes what the question asks.
- Check joins for correct relationships and avoid accidental Cartesian
  products.
- Check aggregation levels and grouping.
- Check filtering conditions and comparison operators.
- Check ordering and LIMIT semantics where applicable.
- Check NULL handling where relevant.
- Check that the query does not unintentionally duplicate or eliminate
  records.

6. RESULT SANITY
- The query should be capable of producing the type of result requested by the
  question.
- Do not add restrictive conditions merely to make the query return a
  particular result.
- Do not assume that a query is correct simply because it is syntactically
  valid.
- A potentially empty result is not automatically incorrect if the question
  and schema
  legitimately allow it.

7. MINIMAL AND FAITHFUL SQL
- Prefer the simplest SQL that correctly answers the question.
- Do not add unnecessary complexity.
- Do not introduce alternative interpretations when the question has a clear
  meaning.
- Do not use information outside the provided context.

--------------------
CONFIDENCE CALIBRATION
--------------------

Assign a confidence score from 0 to 100 representing your confidence that the
  generated
SQL is BOTH:

1. Correctly aligned with the question, and
2. Fully grounded in the provided context.

Confidence must reflect the result of the complete internal validation, not
  merely
confidence that the SQL is syntactically valid.

Use these rules:

- 90-100:
  The question is unambiguous, all required schema elements and values are
    explicitly
  supported, the SQL is clearly aligned with the question, and the SQLite
    syntax and
  logic are highly certain.

- 70-89:
  The SQL is likely correct, but there is minor interpretation or
    schema-related
  uncertainty.

- 40-69:
  There is meaningful uncertainty about intent, values, relationships, or SQL
    logic.

- 1-39:
  Significant uncertainty exists, required information is missing, or the SQL
    may
  contain unsupported assumptions.

- 0:
  The SQL cannot be reliably grounded in the provided context or cannot
    correctly answer
  the question.

IMPORTANT:
- A syntactically valid SQL query does NOT justify high confidence.
- If the SQL contains a hallucinated table, column, relationship, value, or
  unsupported
  assumption, confidence MUST be low.
- If the SQL is not clearly aligned with the question, confidence MUST be low.
- If SQLite compatibility is uncertain, confidence MUST be low.
- Confidence must reflect the weakest critical aspect of the generated SQL
  rather than
  averaging independent checks.

--------------------
FINAL OUTPUT
--------------------

Return EXACTLY ONE valid JSON object.

The JSON MUST contain ONLY these two fields:

{
  "sql_query": "<generated SQLite-compatible SQL query>",
  "confidence": <integer between 0 and 100>
}

Do NOT output:
- explanations
- reasoning
- validation results
- evaluation scores
- additional fields
- markdown
- code fences
- comments outside the JSON

The output must be valid JSON and nothing else.
\end{Verbatim}

\subsection{Stage 2 Prompt (Critic)}
\label{stage_2_prompt}

The critic $\theta_C \neq \theta_G$ receives the question, schema,
generated query, and execution result in one call. Step~1 yields
$v_{\mathrm{sql}}$ (\texttt{evaluation\_score}) and $c_{\mathrm{val}}$
(\texttt{evaluation\_confidence}). Step~2 yields $\tilde{q}$
(\texttt{generated\_question}), and Step~3 yields $a$
(\texttt{alignment\_score}) and $s_{\mathrm{sim}}$
(\texttt{question\_similarity}).

\begin{Verbatim}[fontsize=\small]
You are a strict, conservative, outcome-driven evaluator for Text-to-SQL.

Your task is to determine whether the EXECUTED SQL RESULT correctly and
completely answers the ORIGINAL QUESTION. Do not assume access to a gold
SQL or gold answer.

Inputs:
1. Natural language question
2. Database schema
3. Generated SQL query
4. SQL execution result

Perform the following steps internally and return ONLY the specified JSON.

--------------------------------
STEP 1: STRICT RESULT VALIDATION
--------------------------------

Determine whether the result is a correct and complete answer to the
question. Evaluate the result against the QUESTION, not merely whether
the values appear plausible.

Check all relevant aspects:

1. COLUMN / ROW VALIDITY
- Are the required columns present and relevant?
- Are required rows missing?
- Are extra or invalid rows/columns present?

2. OPERATION / AGGREGATION ALIGNMENT
Verify that the result reflects the operation explicitly requested:
COUNT, SUM, AVG, MIN, MAX, percentage, ratio, rate, difference, total,
most, least, highest, lowest, etc.
A mismatch such as AVG vs SUM or COUNT vs returning individual rows is
incorrect.

3. FILTER / OPERATOR ALIGNMENT
Verify every condition and operator in the question:
>, >=, <, <=, =, !=, BETWEEN, before, after, at least, at most,
exactly, categorical filters, dates, entities, AND/OR, exclusions, etc.
The result must respect the exact semantics and boundaries.

4. ORDER / LIMIT / TOP-K ALIGNMENT
If the question specifies top/bottom, first/latest/oldest, ranking,
sorting, or a number of results, verify both:
- correct ordering/ranking criterion
- correct number of returned records
For example, "top 5 by revenue" requires the five highest-revenue
records, not merely any five records.

5. GROUPING / GRANULARITY
Verify that the result has the requested level of aggregation:
e.g. overall total vs total by company, average by year, results for
each sector, etc.

6. DATE / TIME ALIGNMENT
Verify requested years, dates, periods, before/after semantics, ranges,
latest/earliest requirements, and other temporal constraints.

7. VALUE / RESULT CONSISTENCY
Verify that returned scalar, aggregate, ranking, and derived values are
consistent with the question.

8. COMPLETENESS / EMPTY RESULTS
Verify that all parts of the question are answered. Empty, NULL, or zero
results are not automatically wrong; determine whether they are
consistent with the question and applied conditions.

If ANY substantive requirement is violated, the result is incorrect.

evaluation_score:
- 1 = result is correct AND complete.
- 0 = result is incorrect, incomplete, or violates any substantive
      requirement.

evaluation_confidence:
Confidence (0-100) in the above binary evaluation.
- 90-100: clear evidence and no meaningful ambiguity
- 70-89: strong evidence with minor uncertainty
- 40-69: meaningful uncertainty
- 1-39: substantial uncertainty
- 0: cannot reliably determine correctness

A clearly violated explicit requirement (aggregation, operator, filter,
ordering, LIMIT, grouping, date, etc.) should produce high confidence
that evaluation_score = 0.

--------------------------------
STEP 2: REVERSE QUESTION GENERATION
--------------------------------

Ignore the original question.

Using ONLY the SQL RESULT and schema, generate the most precise question
that the result actually answers.

Do not repair, idealize, or infer missing filters, aggregations,
conditions, ordering, or other requirements.

--------------------------------
STEP 3: ALIGNMENT AND SIMILARITY
--------------------------------

Compare the generated question with the original question.

alignment_score (0-100) measures how well the RESULT itself answers the
original question, considering:
- requested information
- filters and operators
- aggregation
- ordering
- LIMIT/top-k
- grouping/granularity
- dates/time
- completeness

question_similarity (0-100) measures semantic similarity between the
ORIGINAL QUESTION and the GENERATED QUESTION.

Important:
- A result can be topically related but operationally incorrect.
- Do not give high alignment merely because the returned data is related.
- If evaluation_score = 0, alignment_score should normally be <= 30
  for a substantive correctness failure.
- If the result only partially answers the question, alignment_score
  should normally be <= 50.
- If evaluation_score = 0, question_similarity MUST NOT exceed 60.

--------------------------------
OUTPUT
--------------------------------

Return EXACTLY ONE valid JSON object with these five fields:

{
  "evaluation_score": <0 or 1>,
  "evaluation_confidence": <integer between 0 and 100>,
  "alignment_score": <number between 0 and 100>,
  "generated_question": "<generated question>",
  "question_similarity": <number between 0 and 100>
}

No markdown.
No explanations.
No reasoning.
No additional fields.

--------------------------------
INPUT
--------------------------------

Original Question:
<INSERT_QUESTION_HERE>

Database Schema:
<INSERT_DB_SCHEMA_HERE>

Generated SQL Query:
<INSERT_SQL_QUERY_HERE>

SQL Result:
<INSERT_RECORDS_HERE>
\end{Verbatim}

\newpage

\section{Full Feature List}
\label{appendix:feature_list}

Table~\ref{tab:feature_list} lists all 41 features of $\mathbf{x}$ with
the identifiers used in our code. All token statistics use the
generator's probabilities $\pi_t$ of the tokens of $\hat{s}$, from the
first \textsc{select} token to the terminating semicolon.
\emph{Keyword} tokens are upper-case SQL keywords, and
\emph{non-keyword} tokens are the rest (identifiers, literals,
operators). The family column gives the grouping used in the ablation
(Table~\ref{tab:ablation}).

\begin{table}[H]
\centering
\scriptsize
\caption{All 41 features. Core features appear in the main text
(Table~\ref{tab:features}) and in the density plots of
Appendix~\ref{appendix:feature_distribution}.}
\label{tab:feature_list}
\setlength{\tabcolsep}{3pt}
\renewcommand{\arraystretch}{1.1}
\begin{tabular}{l l l p{5.3cm} c}
\toprule
\textbf{Family} & \textbf{Symbol} & \textbf{Code name} & \textbf{Definition} & \textbf{Core} \\
\midrule
Verbalized & $c_{\mathrm{gen}}$ & \texttt{Generation\_conf.} & Generator's verbalized confidence, 0--100 (Stage 1) & \checkmark \\
 & $v_{\mathrm{sql}}$ & \texttt{Evaluation\_Score} & Critic's binary verdict on the result (Stage 2) & \checkmark \\
 & $c_{\mathrm{val}}$ & \texttt{Evaluation\_Conf.} & Critic's confidence in its verdict, 0--100 (Stage 2) & \checkmark \\
\midrule
Alignment & $a$ & \texttt{alignment\_score} & How well the result answers $q$, 0--100 (Stage 2) & \checkmark \\
 & $s_{\mathrm{sim}}$ & \texttt{question\_similarity} & Similarity of $q$ and reconstructed $\tilde{q}$, 0--100 (Stage 2) & \checkmark \\
\midrule
Logit & $\mu_{\mathrm{geom}}$ & \texttt{geo\_sql\_all} & Geometric mean of $\pi_t$, all tokens & \checkmark \\
($P_{\mathrm{feats}}$) & $\mathrm{FTC}$ & \texttt{ftc\_sql\_all} & Product of $\pi_t$, all tokens & \checkmark \\
 & $\mu_{\mathrm{arith}}$ & \texttt{mean\_sql\_all} & Arithmetic mean of $\pi_t$, all tokens & \checkmark \\
 & $\pi_{\min}$ & \texttt{min\_sql\_all} & Minimum $\pi_t$, all tokens & \checkmark \\
 & $H$ & \texttt{entrop\_sql\_all} & $-\sum_t \pi_t \log \pi_t$, all tokens & \checkmark \\
 & $\sigma^2$ & \texttt{sql\_token\_variance} & Variance of $\pi_t$ & \checkmark \\
 & $\ell_{\mathrm{CLM}}$ & \texttt{clm} & $-\sum_t \log \pi_t$ ($= -\log \mathrm{FTC}$) & \checkmark \\
 & & \texttt{*\_sql\_clause} & \texttt{geo}, \texttt{ftc}, \texttt{mean}, \texttt{min}, \texttt{entrop} over keyword tokens (5 features) & \\
 & & \texttt{*\_non\_sql} & The same five over non-keyword tokens (5 features) & \\
 & & \texttt{min} & Minimum $\pi_t$ after dropping whitespace tokens (near-duplicate of $\pi_{\min}$) & \\
 & & \texttt{sql\_token\_margin} & Probability range $\max_t \pi_t - \min_t \pi_t$ & \\
 & & \texttt{bin\_1}--\texttt{bin\_13} & Histogram over 13 probability bins, $[0,0.3)$, $[0.3,0.4)$, $[0.4,0.5)$, then width-0.05 bins to 1. Each value is the mean of the min--max-normalised probabilities of the tokens in the bin (0 if empty) (13 features) & \\
\midrule
SQL-structural & $\mathrm{SAC}_{\mathrm{geom}}$ & \texttt{sac} & Geometric mean over clause groups of the within-group mean $\pi_t$ (Eq.~\ref{eq:sac}) & \checkmark \\
($S_{\mathrm{feats}}$) & $\mathrm{SLC}$ & \texttt{slc} & Geometric mean of $\pi_t$ over schema-element tokens (Eq.~\ref{eq:sac}) & \checkmark \\
 & & \texttt{schema\_geo\_mean} & Geometric mean of $\pi_t$ over tokens matching schema table names & \\
 & & \texttt{c3s} & Weakest clause geometric mean $\times\, e^{-\mathrm{Var}_{\text{inter}}} \times e^{-\overline{\mathrm{Var}}_{\text{intra}}}$ over clause groups & \\
\midrule
\multicolumn{3}{l}{\textbf{Total}: 5 LLM-derived + 32 logit + 4 structural} & & \textbf{41} (14 core) \\
\bottomrule
\end{tabular}
\end{table}

Some features are constant in some configurations and are then never
used by the forest. For example, \texttt{bin\_1} (probabilities below
0.3) is always empty for generator GPT-OSS-120B.

\section{Feature Distributions Across Generator Backbones}
\label{appendix:feature_distribution}

\begin{figure*}[ht!]
\centering
\includegraphics[width=1.0\textwidth]{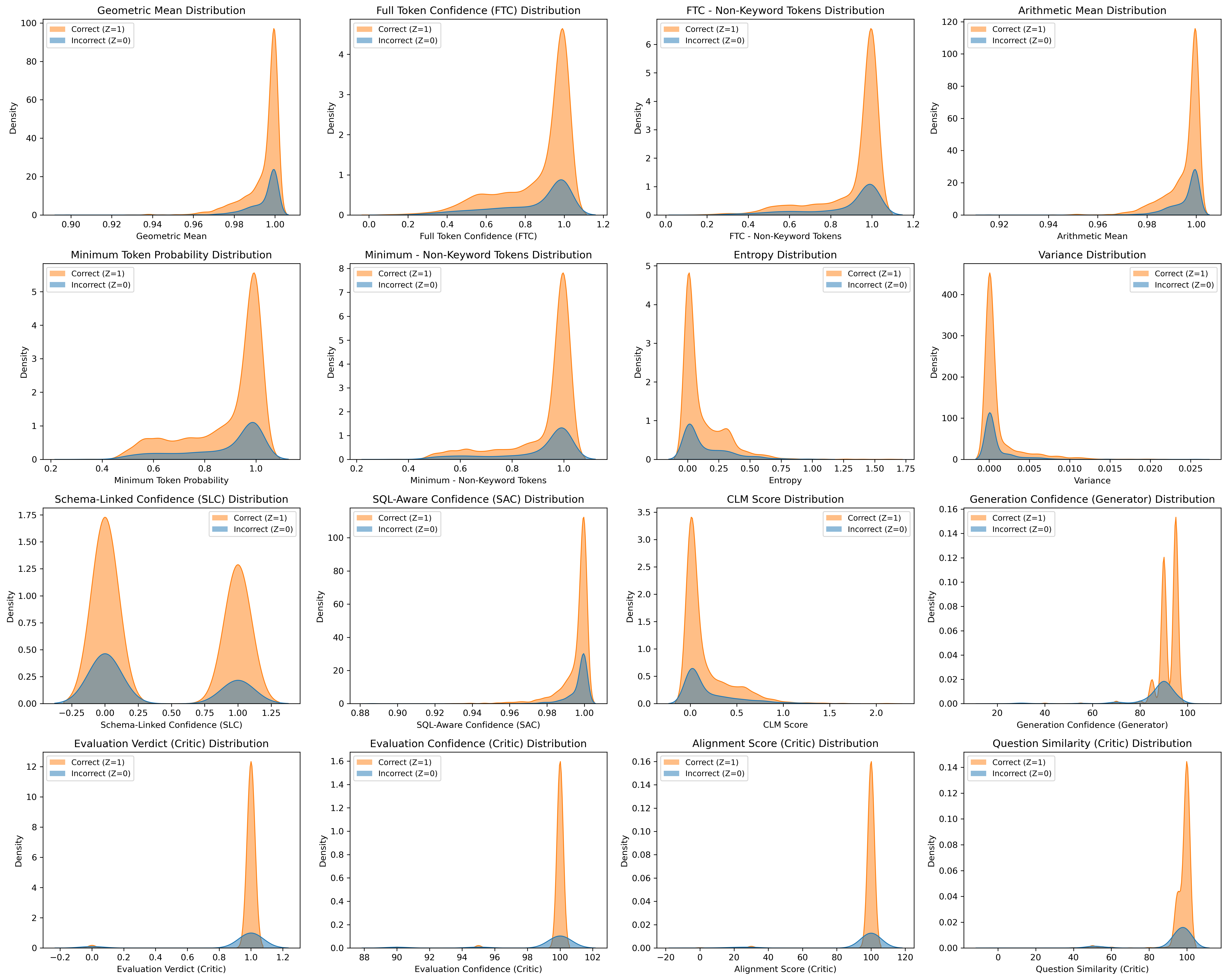}
\caption[KDEs of confidence features]{Kernel density estimates of the 14 core features and two non-keyword variants for generator GPT-OSS-120B with the Gemma-4-31B critic, separated by execution outcome (orange: correct, blue: incorrect). Feature selection is described at the start of Appendix~\ref{appendix:feature_distribution}.}
\label{fig:logit_features_GPT}
\end{figure*}

Figures~\ref{fig:logit_features_GPT}--\ref{fig:logit_features_Qwen} show
kernel density estimates for 16 features per generator backbone: the
14 core features of Table~\ref{tab:feature_list}, and FTC and the
minimum probability computed on non-keyword tokens
(\texttt{ftc\_non\_sql}, \texttt{min\_non\_sql}). The two non-keyword
variants are included because identifiers and literals are where
schema-grounding errors appear. We omit the other 25 features for
readability. They recompute a core statistic on a token subset, are
histogram bins, are the near-duplicate \texttt{min} or the range
\texttt{sql\_token\_margin}, or are the structural variants
\texttt{schema\_geo\_mean} and \texttt{c3s}. In every panel, correct
queries ($Z{=}1$) are shown in orange and incorrect queries ($Z{=}0$)
in blue. Densities are normalised jointly, so each curve's area is
proportional to its class size. The script \texttt{plot\_feature\_kde.py}
reproduces the figures from the saved feature tables.

\newpage
\subsection{GPT-OSS-120B}

\newpage
\subsection{Olmo-3-32B-Think}
\begin{figure*}[ht!]
\centering
\includegraphics[width=1.0\textwidth]{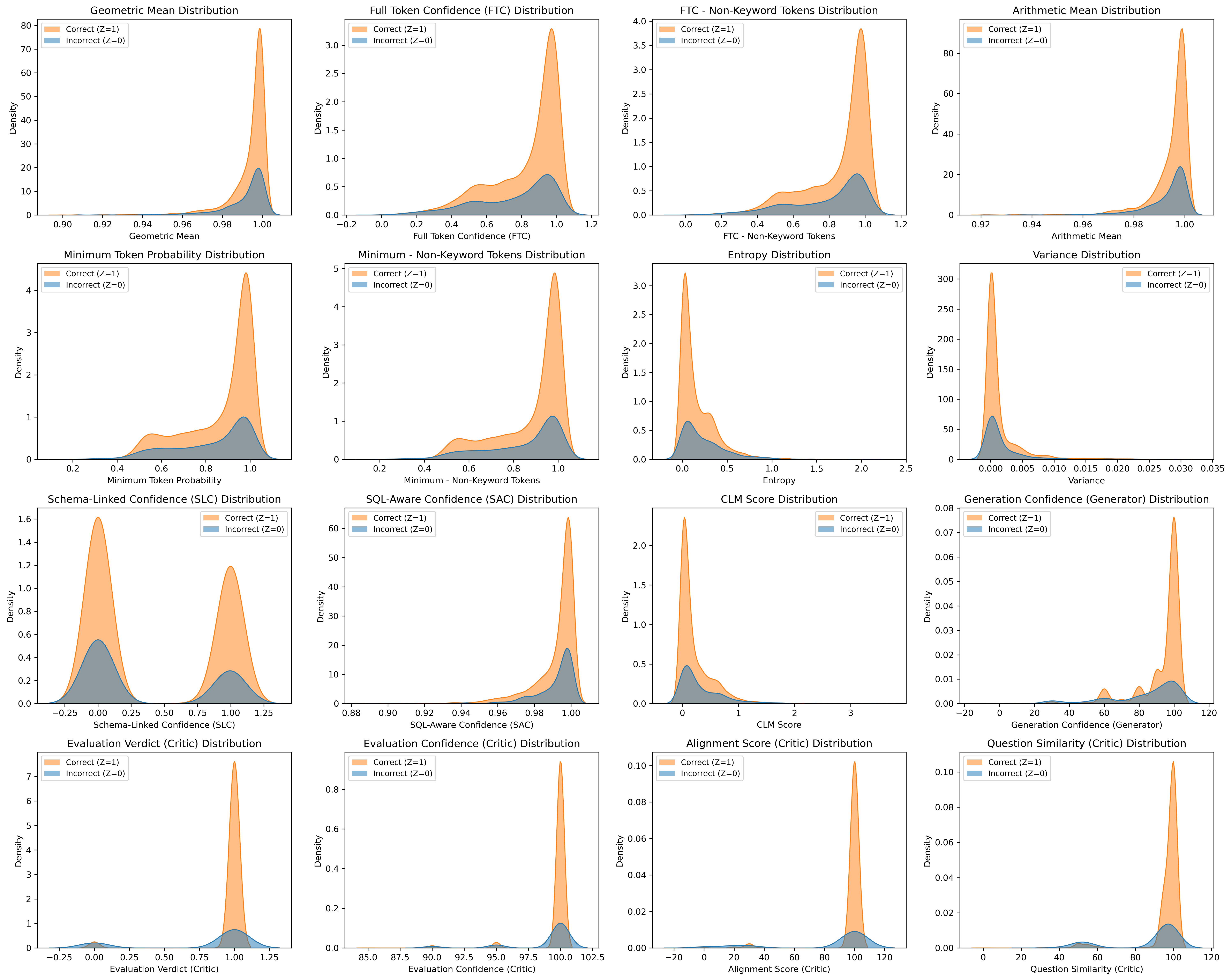}
\caption[KDEs of confidence features]{Kernel density estimates of the 14 core features and two non-keyword variants for generator Olmo-3-32B-Think with the Gemma-4-31B critic, separated by execution outcome (orange: correct, blue: incorrect). Feature selection is described at the start of Appendix~\ref{appendix:feature_distribution}.}
\label{fig:logit_features_olmo}
\end{figure*}

\newpage
\subsection{Llama-3.3-Nemotron-Super-49B-v1.5}
\begin{figure*}[ht!]
\centering
\includegraphics[width=1.0\textwidth]{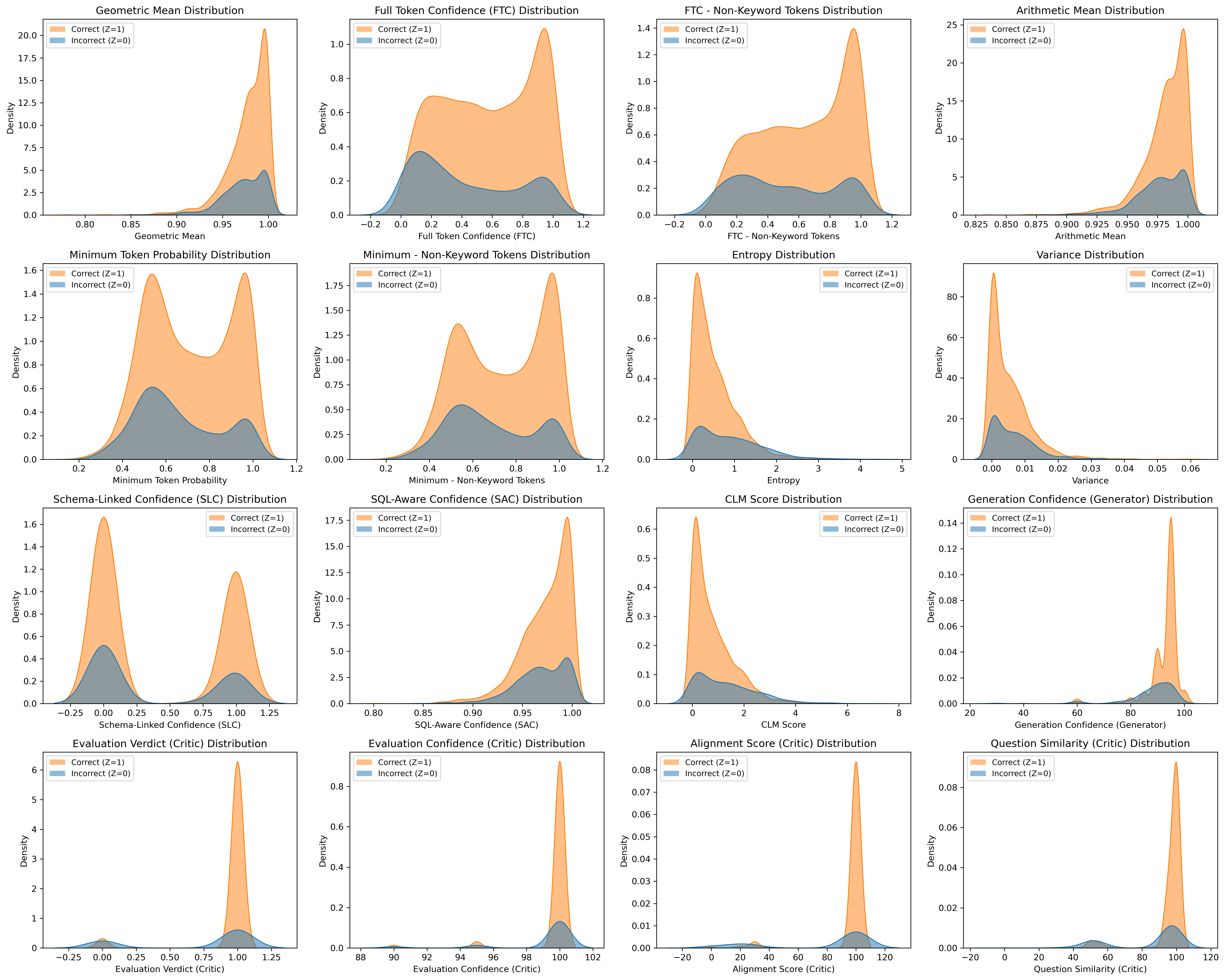}
\caption[KDEs of confidence features]{Kernel density estimates of the 14 core features and two non-keyword variants for generator Llama-3.3-Nemotron-Super-49B-v1.5 with the Gemma-4-31B critic, separated by execution outcome (orange: correct, blue: incorrect). Feature selection is described at the start of Appendix~\ref{appendix:feature_distribution}.}
\label{fig:logit_features_Nemotron}
\end{figure*}

\newpage
\subsection{Qwen2.5-72B-Instruct}
\begin{figure*}[ht!]
\centering
\includegraphics[width=1.0\textwidth]{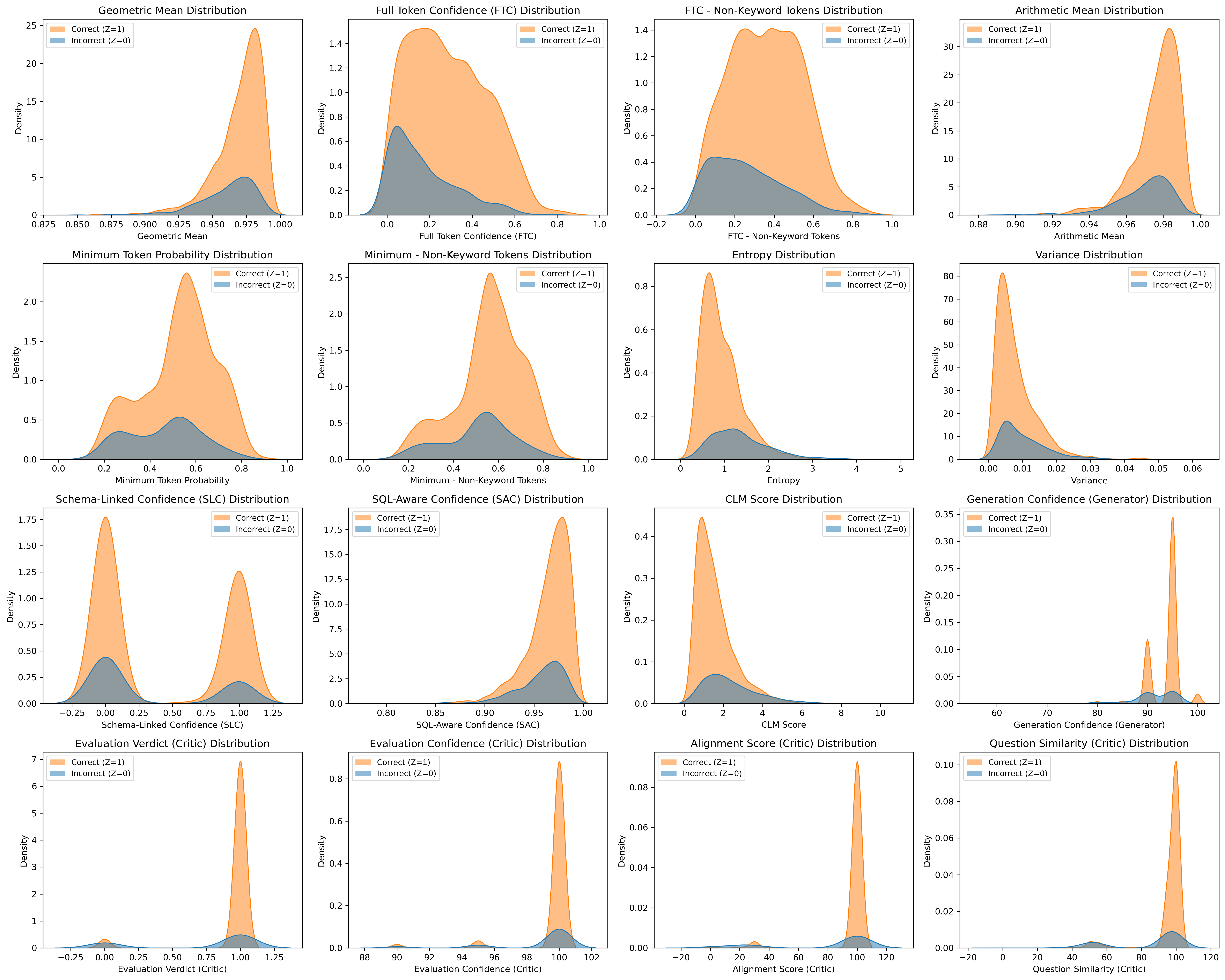}
\caption[KDEs of confidence features]{Kernel density estimates of the 14 core features and two non-keyword variants for generator Qwen2.5-72B-Instruct with the Gemma-4-31B critic, separated by execution outcome (orange: correct, blue: incorrect). Feature selection is described at the start of Appendix~\ref{appendix:feature_distribution}.}
\label{fig:logit_features_Qwen}
\end{figure*}
\clearpage
\section{Random Forest Training}
\label{appendix:random_forest}

\textbf{Hyperparameters.} For each configuration, hyperparameters are
chosen by randomized search over 40 configurations with 5-fold
cross-validation on the training split, maximising ROC-AUC. Bootstrap
sampling and \texttt{class\_weight=balanced\_subsample} are always used,
and out-of-bag estimation is enabled. Table~\ref{tab:rf_params} lists
the selected values.

\begin{table}[H]
\centering
\small
\caption{Selected Random Forest hyperparameters per configuration.
$^\ast$Generation run of the four-backbone study (Section~\ref{sec:setup}).}
\label{tab:rf_params}
\resizebox{\linewidth}{!}{%
\begin{tabular}{lccccc}
\toprule
Configuration & Trees & Max depth & Min split & Min leaf & Max features \\
\midrule
Generator GPT-OSS-120B, critic Gemma-4-31B & 400 & 10 & 5 & 10 & \texttt{sqrt} \\
Generator GPT-OSS-120B, critic Qwen3.8-27B & 400 & 10 & 5 & 10 & \texttt{sqrt} \\
Generator GPT-OSS-120B, critic DeepSeek-V4.1-Flash & 400 & 10 & 5 & 10 & \texttt{sqrt} \\
Generator GPT-OSS-120B, critic Nemotron-3-Super & 200 & 50 & 10 & 10 & all \\
Generator GPT-OSS-120B$^\ast$, critic Gemma-4-31B & 800 & 10 & 10 & 10 & all \\
Generator Olmo-3-32B, critic Gemma-4-31B & 400 & 15 & 5 & 10 & \texttt{log2} \\
Generator Qwen2.5-72B, critic Gemma-4-31B & 200 & 20 & 2 & 1 & \texttt{log2} \\
Generator Llama-49B, critic Gemma-4-31B & 400 & 10 & 5 & 10 & \texttt{sqrt} \\
\bottomrule
\end{tabular}}
\end{table}

\textbf{Per-configuration training.} A separate forest is trained for
each configuration. Its training set contains the feature vectors
computed from that configuration's outputs on the Spider training split, with labels
$Z_i$ obtained by execution matching against gold SQL. We train per
backbone because logit-derived features ($P_{\mathrm{feats}}$,
$S_{\mathrm{feats}}$) have backbone-specific distributions. Token
probability ranges, entropy scales, and SAC values differ across model
families and sizes
(Figures~\ref{fig:logit_features_GPT}--\ref{fig:logit_features_Qwen}).
For zero-shot transfer (Section~\ref{sec:transfer}), the GPT-OSS-120B$^\ast$
forest and calibration scores (Table~\ref{tab:rf_params}) are applied to GPT-OSS-120B outputs on
BIRD. For NL2SQL-BUGs, whose SQL is given, a forest trained on the four
critic scores is used. No target-domain labels are used at any stage.

\textbf{Conformal calibration.} Calibration scores $\eta_i = -\log p_i$
are computed on calibration queries with $Z_i = 1$ only. At test time,
Eq.~\ref{eq:conformal} is evaluated with ties counted in the numerator
($\eta_i \geq \eta$). Counting ties this way makes $\hat{c}$ larger and
the FAR bound conservative. Using $1 - p$ instead of $-\log p$ as the
nonconformity score gives the same $\hat{c}$, because the two are
monotone transforms of each other.

\section{Proofs}
\label{appendix:proofs}

\textbf{Proposition~\ref{prop:conformal}.} Let $\eta_1,\dots,\eta_{m_1}$
be the scores of the correct calibration queries, and $\eta_{m_1+1}$
the score of a correct test query. Under exchangeability, conditional
on the trained estimator $p$, the rank of $\eta_{m_1+1}$ among the
$m_1+1$ scores is uniform on $\{1,\dots,m_1+1\}$. Ties are counted in
the numerator of Eq.~\ref{eq:conformal}, which can only increase
$\hat{c}$. Hence $\hat{c} = (1 + |\{i \leq m_1 : \eta_i \geq
\eta_{m_1+1}\}|)/(m_1+1)$ satisfies $\mathbb{P}(\hat{c} \leq t \mid Z=1)
\leq \lfloor t(m_1+1) \rfloor/(m_1+1) \leq t$. Taking $t = 1-\alpha$
gives $\mathrm{FAR}(1-\alpha) = \mathbb{P}(\hat{c} < 1-\alpha \mid Z=1)
\leq 1-\alpha$~\citep[Prop.~3]{angelopoulos2023conformal}.

\textbf{Proposition~\ref{prop:risk}.} Fix $\lambda \in \Lambda$. Since
$p$ is trained on data disjoint from $\mathcal{C}$ and the calibration
queries are i.i.d., the queries with $p_i \geq \lambda$ are, conditional
on their number $n(\lambda)$, i.i.d.\ draws from the distribution of
queries with $p \geq \lambda$. Their error count therefore satisfies
$k(\lambda) \mid n(\lambda) \sim \mathrm{Bin}(n(\lambda),
\mathrm{Risk}(\lambda))$. Under $H_\lambda: \mathrm{Risk}(\lambda) >
\varepsilon$, binomial stochastic ordering gives
$\mathbb{P}(u_\lambda \leq s) \leq s$ for all $s$, so $u_\lambda$ is a
valid p-value. The rule certifies only thresholds with $u_\lambda <
\delta/|\Lambda|$. By the union bound, the probability of certifying
any $\lambda$ with $\mathrm{Risk}(\lambda) > \varepsilon$ is at most
$\delta$. On the complementary event, $\hat{\lambda}$ satisfies
$\mathrm{Risk}(\hat{\lambda}) \leq \varepsilon$, or no threshold is
certified and the system abstains on every query. This is the Learn
then Test guarantee~\citep{angelopoulos2022learn} with the Bonferroni
correction~\citep[Thm.~A.1]{angelopoulos2023conformal}.

\section{Evaluation Metrics}
\label{eval_metric}

All metrics are computed on the test split. For selective metrics, a
query is \emph{accepted} if its score passes the rule's threshold
($\hat{c} \geq 1-\alpha$ for the conformal rule, $p \geq
\hat{\lambda}$ for the risk-controlled rule) and \emph{abstained}
otherwise. We treat acceptance as the positive
prediction and write TP (correct, accepted), FP (incorrect, accepted),
FN (correct, abstained), and TN (incorrect, abstained).

\begin{table}[H]
  \centering
  \small
  \caption{Evaluation metrics. Risk is bounded by
  Proposition~\ref{prop:risk} (risk-controlled rule), and FAR by
  Proposition~\ref{prop:conformal} (conformal rule).}
  \label{tab:eval_metric}
  \setlength{\tabcolsep}{4pt}
  \renewcommand{\arraystretch}{1.2}
  \begin{tabular}{p{2.4cm} p{9.0cm} p{1.4cm}}
    \toprule
    \textbf{Metric} & \textbf{Definition} & \textbf{Better} \\
    \midrule
    ECE & Weighted mean gap between confidence and empirical accuracy over 10 equal-width bins. & $\downarrow$ \\
    ACE & As ECE, with equal-mass bins. & $\downarrow$ \\
    Brier & Mean squared error between confidence and $Z$. & $\downarrow$ \\
    AUROC & Probability that a random correct query receives a higher score than a random incorrect one. & $\uparrow$ \\
    AUARC & Area under the accuracy--rejection curve: accuracy of the retained queries as the rejection rate varies from 0 to 1. Primary metric. & $\uparrow$ \\
    Prec./Recall/F1 (Table~\ref{tab:full_multillm}) & Classification of $Z{=}1$ by the forest at $p \geq 0.5$. & $\uparrow$ \\
    Error-detection Prec./Recall/F1 (BIRD, NL2SQL-BUGs) & Classification of $Z{=}0$ by flagging queries with $\hat{c} < 1-\alpha$. & $\uparrow$ \\
    Coverage & Fraction of queries accepted, $(\mathrm{TP}+\mathrm{FP})/n$. Empirical. & --- \\
    Risk & Error rate among accepted queries, $\mathrm{FP}/(\mathrm{TP}+\mathrm{FP})$. Bounded by $\varepsilon$ with probability $1-\delta$ for the risk-controlled rule under i.i.d.\ data (Proposition~\ref{prop:risk}). & $\downarrow$ \\
    CondAcc & Accuracy among accepted queries, $1-\mathrm{Risk} = \mathrm{TP}/(\mathrm{TP}+\mathrm{FP})$. Identical to SSC($|S|{=}1$). Empirical. & $\uparrow$ \\
    TAR & True abstention rate, $\mathrm{TN}/(\mathrm{TN}+\mathrm{FP})$: fraction of incorrect queries withheld. Empirical. & $\uparrow$ \\
    FAR & False abstention rate, $\mathrm{FN}/(\mathrm{FN}+\mathrm{TP})$: fraction of correct queries withheld. Bounded by $1-\alpha$ in expectation under exchangeability. & $\downarrow$ \\
    SSC($|S|{=}0$) & Accuracy among abstained queries, $\mathrm{FN}/(\mathrm{FN}+\mathrm{TN})$. & --- \\
    \bottomrule
  \end{tabular}
\end{table}

\section{Additional Experimental Results}

\subsection{Model Details}
\label{appendix:model_details}

\begin{table}[H]
  \caption{Language models used in our experiments. The first four
  serve as generators. Gemma-4-31B-it is the default critic, and
  Qwen3.8-27B, DeepSeek-V4.1-Flash, and Nemotron-3-Super are additional
  critics. Qwen2.5-72B-Instruct also serves as a critic. The main configuration pairs generator
  GPT-OSS-120B with critic Gemma-4-31B-it.}
  \label{tab:models_used}
  \centering
  \small
  \begin{tabularx}{\textwidth}{L{3cm} c Y}
    \toprule
    Model & Params (B) & HuggingFace model ID \\
    \midrule
    GPT-OSS-120B & 120 & \path{openai/gpt-oss-120b} \\
    Olmo-3-32B-Think & 32 & \path{allenai/Olmo-3-32B-Think} \\
    Llama-3.3-Nemotron-Super-49B-v1.5 & 49 & \path{nvidia/Llama-3_3-Nemotron-Super-49B-v1_5} \\
    Qwen2.5-72B-Instruct & 72 & \path{Qwen/Qwen2.5-72B-Instruct} \\
    Gemma-4-31B-it (critic) & 31 & \path{google/gemma-4-31B-it} \\
    Qwen3.8-27B (critic) & 27 & \path{Qwen/Qwen3.8-27B} \\
    DeepSeek-V4.1-Flash (critic) & --- & \path{deepseek-ai/DeepSeek-V4.1-Flash} \\
    Nemotron-3-Super (critic) & 120 & \path{nvidia/NVIDIA-Nemotron-3-Super-120B-A12B-BF16} \\
    \bottomrule
  \end{tabularx}
\end{table}

\subsection{Full Results for Four Generator Backbones}
\label{appendix:full_multillm}

Table~\ref{tab:full_multillm} reports the four generator backbones, each
with the Gemma-4-31B critic. The confusion matrices at $\alpha{=}0.1$
reproduce the reported TAR, FAR, and CondAcc. FAR and CondAcc use
different denominators. For Olmo-3-32B, FAR is computed over the
470 correct queries (416 withheld, 0.885), whereas CondAcc is computed
over the 56 accepted queries (54 correct, 0.964).

\begin{table}[H]
\centering
\small
\caption{Four generator backbones with the Gemma-4-31B critic on the
Spider test split. Selective metrics use the conformal rule at
$\alpha{=}0.1$. The GPT-OSS-120B column uses the generation run of the
four-backbone study, not the main run of Table~\ref{tab:critic}.}
\label{tab:full_multillm}
\setlength{\tabcolsep}{5pt}
\begin{tabular}{lcccc}
\toprule
\textbf{Metric} & \textbf{GPT-OSS-120B} & \textbf{Olmo-3-32B} & \textbf{Qwen2.5-72B} & \textbf{Llama-49B} \\
\midrule
Correct / incorrect queries & 499 / 137 & 470 / 166 & 503 / 133 & 475 / 160 \\
Base accuracy $\rho$ & 0.785 & 0.739 & 0.791 & 0.748 \\
\midrule
ECE    & 0.133 & 0.126 & 0.045 & 0.113 \\
ACE    & 0.133 & 0.125 & 0.041 & 0.112 \\
Brier  & 0.163 & 0.158 & 0.137 & 0.153 \\
AUROC  & 0.791 & 0.816 & 0.765 & 0.814 \\
AUARC  & 0.918 & 0.895 & 0.898 & 0.902 \\
Precision ($Z{=}1$, $p\geq 0.5$) & 0.889 & 0.847 & 0.815 & 0.854 \\
Recall ($Z{=}1$, $p\geq 0.5$)    & 0.804 & 0.851 & 0.974 & 0.848 \\
F1 ($Z{=}1$, $p\geq 0.5$)        & 0.844 & 0.849 & 0.888 & 0.851 \\
\midrule
TP / FP / FN / TN & 51 / 1 / 448 / 136 & 54 / 2 / 416 / 164 & 50 / 3 / 453 / 130 & 52 / 1 / 423 / 159 \\
Coverage & 0.082 & 0.088 & 0.083 & 0.083 \\
Risk     & 0.019 & 0.036 & 0.057 & 0.019 \\
CondAcc $=1-$Risk & 0.981 & 0.964 & 0.943 & 0.981 \\
TAR      & 0.993 & 0.988 & 0.977 & 0.994 \\
FAR      & 0.898 & 0.885 & 0.901 & 0.891 \\
SSC ($|S|{=}0$) & 0.767 & 0.717 & 0.777 & 0.727 \\
\bottomrule
\end{tabular}
\end{table}

\subsection{Operating Points}
\label{appendix:operating_points}

\textbf{Risk-controlled rule on the reported split.} Table~\ref{tab:ltt_single}
applies the risk-controlled rule to the split behind
Tables~\ref{tab:critic} and~\ref{tab:full_multillm}. On this split,
every certified threshold keeps the test risk below $\varepsilon$.
Critic rows use generator GPT-OSS-120B; generator rows use the Gemma
critic and the generation runs of the four-backbone study.

\begin{table}[H]
\centering
\small
\caption{Risk-controlled rule ($\delta{=}0.1$) on the reported Spider
split. $\hat{\lambda}$: certified threshold on $p$ (``---'': none
certified, so the system abstains on every query). Cov.\ and Risk in \%.}
\label{tab:ltt_single}
\setlength{\tabcolsep}{3.5pt}
\begin{tabular}{l ccc ccc ccc}
\toprule
 & \multicolumn{3}{c}{$\varepsilon = 0.10$} & \multicolumn{3}{c}{$\varepsilon = 0.15$} & \multicolumn{3}{c}{$\varepsilon = 0.20$} \\
\cmidrule(lr){2-4}\cmidrule(lr){5-7}\cmidrule(lr){8-10}
 & $\hat{\lambda}$ & Cov. & Risk & $\hat{\lambda}$ & Cov. & Risk & $\hat{\lambda}$ & Cov. & Risk \\
\midrule
Critic: Gemma-4-31B & --- & 0.0 & --- & 0.58 & 54.4 & 8.4 & 0.44 & 76.4 & 13.4 \\
Critic: Qwen3.8-27B & 0.68 & 44.3 & 5.7 & 0.50 & 67.5 & 9.6 & 0.40 & 79.7 & 12.4 \\
Critic: DeepSeek-V4.1-Flash & 0.77 & 29.4 & 2.7 & 0.59 & 54.7 & 8.9 & 0.41 & 81.0 & 14.4 \\
Critic: Nemotron-3-Super & --- & 0.0 & --- & 0.65 & 47.3 & 7.3 & 0.46 & 71.1 & 13.1 \\
Gen.: GPT-OSS-120B & 0.91 & 20.1 & 0.8 & 0.67 & 47.6 & 5.9 & 0.56 & 61.6 & 8.7 \\
Gen.: Olmo-3-32B & --- & 0.0 & --- & 0.77 & 26.4 & 3.6 & 0.51 & 73.0 & 14.7 \\
Gen.: Qwen2.5-72B & 0.87 & 42.3 & 8.6 & 0.74 & 72.5 & 10.8 & 0.60 & 89.6 & 16.5 \\
Gen.: Llama-49B & --- & 0.0 & --- & 0.67 & 50.9 & 9.3 & 0.57 & 65.8 & 11.7 \\
\bottomrule
\end{tabular}
\end{table}

\textbf{Conformal rule.} Table~\ref{tab:operating_points} sweeps the
operating level $\alpha$ of the conformal rule. By
Eq.~\ref{eq:identity}, $\alpha$ fixes the accepted fraction of correct
queries at about $\alpha$, and the estimator determines how many
incorrect queries are accepted alongside them ($1-\mathrm{TAR}$). With
the Gemma critic, coverage rises from 7.9\% to 100\%. Risk stays between 5.0\% and
6.8\% up to $\alpha{=}0.6$ and then rises to the no-abstention error of
22.6\%. On this single
split, FAR exceeds its bound by up to 0.049 (at $\alpha{=}0.5$). All levels
share one calibration draw, so their deviations are correlated. The
bound holds in expectation over draws. The 10-seed mean FAR at
$\alpha{=}0.1$ is $0.907 \pm 0.016$, within one standard deviation of
the bound (Appendix~\ref{appendix:seeds}).

\begin{table}[H]
\centering
\small
\caption{Coverage--risk trade-off of the conformal rule
(Proposition~\ref{prop:conformal}) on the Spider test split. Critic rows
use generator GPT-OSS-120B. The last rows average the four generator
backbones of Table~\ref{tab:full_multillm} (Gemma critic).}
\label{tab:operating_points}
\setlength{\tabcolsep}{2.6pt}
\begin{tabular}{llcccccccccc}
\toprule
 & $\alpha$ & 0.1 & 0.2 & 0.3 & 0.4 & 0.5 & 0.6 & 0.7 & 0.8 & 0.9 & 1.0 \\
\midrule
 & FAR bound $1-\alpha$ & 0.9 & 0.8 & 0.7 & 0.6 & 0.5 & 0.4 & 0.3 & 0.2 & 0.1 & 0.0 \\
\midrule
\multirow{3}{*}{\shortstack[l]{Critic:\\Gemma-4-31B}}
 & FAR       & .904 & .813 & .728 & .640 & .549 & .445 & .319 & .181 & .075 & .000 \\
 & Coverage (\%) & 7.9 & 15.3 & 22.2 & 29.6 & 37.3 & 46.1 & 58.2 & 72.3 & 85.7 & 100 \\
 & Risk (\%) & 6.0 & 5.2 & 5.0 & 5.9 & 6.3 & 6.8 & 9.5 & 12.4 & 16.5 & 22.6 \\
\midrule
\multirow{2}{*}{\shortstack[l]{Critic:\\Qwen3.8-27B}}
 & Coverage (\%) & 6.3 & 14.2 & 21.2 & 26.9 & 35.4 & 46.7 & 58.3 & 69.2 & 80.0 & 100 \\
 & Risk (\%) & 0.0 & 3.3 & 2.2 & 2.3 & 4.0 & 5.7 & 6.7 & 9.3 & 12.6 & 22.6 \\
\midrule
\multirow{2}{*}{\shortstack[l]{Four generators\\(average)}}
 & Coverage (\%) & 8.4 & 15.8 & 23.4 & 31.7 & 39.8 & 51.2 & 61.9 & 72.9 & 84.2 & 100 \\
 & Risk (\%) & 3.3 & 3.3 & 4.1 & 4.7 & 6.3 & 8.8 & 10.5 & 13.1 & 16.9 & 23.4 \\
\bottomrule
\end{tabular}
\end{table}

Table~\ref{tab:critic_alpha} gives the confusion matrices behind the
critic rows of Table~\ref{tab:operating_points}, computed from the
test-split predictions of each configuration (492 correct and 144
incorrect queries).

\begin{table}[H]
\centering
\small
\caption{Accept/abstain confusion matrices across $\alpha$ (generator
GPT-OSS-120B, Spider test split, $n{=}636$). Positive = accepted.}
\label{tab:critic_alpha}
\setlength{\tabcolsep}{3.5pt}
\begin{tabular}{r|rrrrcc|rrrrcc}
\toprule
 & \multicolumn{6}{c|}{Critic: Gemma-4-31B} & \multicolumn{6}{c}{Critic: Qwen3.8-27B} \\
$\alpha$ & TP & FP & FN & TN & TAR & FAR & TP & FP & FN & TN & TAR & FAR \\
\midrule
0.1 & 47 & 3 & 445 & 141 & .979 & .904 & 40 & 0 & 452 & 144 & 1.000 & .919 \\
0.2 & 92 & 5 & 400 & 139 & .965 & .813 & 87 & 3 & 405 & 141 & .979 & .823 \\
0.3 & 134 & 7 & 358 & 137 & .951 & .728 & 132 & 3 & 360 & 141 & .979 & .732 \\
0.4 & 177 & 11 & 315 & 133 & .924 & .640 & 167 & 4 & 325 & 140 & .972 & .661 \\
0.5 & 222 & 15 & 270 & 129 & .896 & .549 & 216 & 9 & 276 & 135 & .938 & .561 \\
0.6 & 273 & 20 & 219 & 124 & .861 & .445 & 280 & 17 & 212 & 127 & .882 & .431 \\
0.7 & 335 & 35 & 157 & 109 & .757 & .319 & 346 & 25 & 146 & 119 & .826 & .297 \\
0.8 & 403 & 57 & 89 & 87 & .604 & .181 & 399 & 41 & 93 & 103 & .715 & .189 \\
0.9 & 455 & 90 & 37 & 54 & .375 & .075 & 445 & 64 & 47 & 80 & .556 & .096 \\
1.0 & 492 & 144 & 0 & 0 & .000 & .000 & 492 & 144 & 0 & 0 & .000 & .000 \\
\bottomrule
\end{tabular}
\end{table}

\begin{figure}[H]
    \centering
    \includegraphics[width=0.45\textwidth]{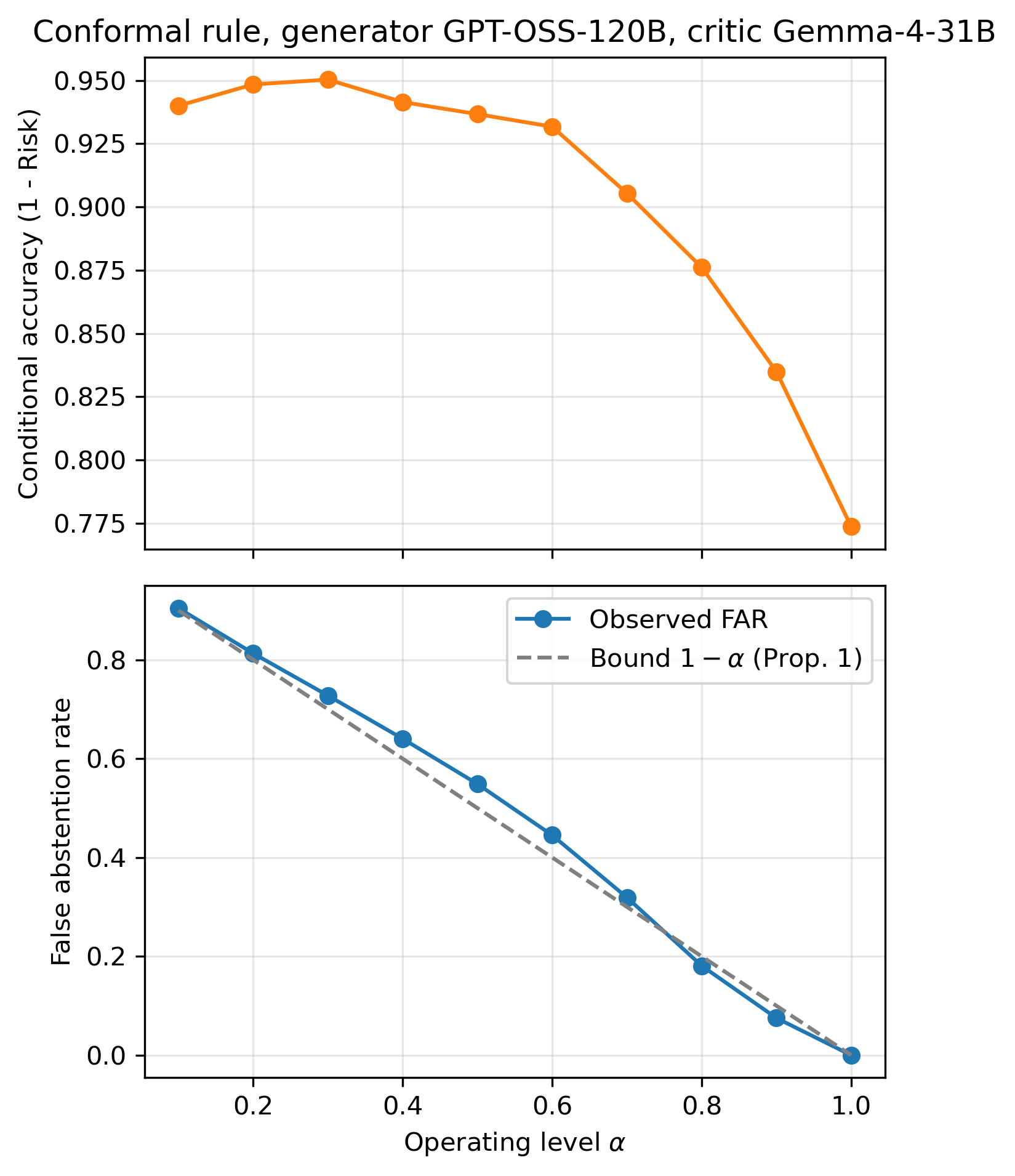}
    \caption{Conditional accuracy (top) and FAR (bottom) of the conformal
    rule across operating levels $\alpha$ (generator GPT-OSS-120B, critic
    Gemma-4-31B). The dashed line is the FAR bound $1-\alpha$ of
    Proposition~\ref{prop:conformal}.}
    \label{fig:conformal_ecr_far}
\end{figure}

\textbf{Generator Olmo-3-32B.} Table~\ref{tab:olmo_alpha} gives the
confusion matrix at each operating level for generator Olmo-3-32B with
the Gemma critic. Recall of correct queries, $1-\mathrm{FAR}$, stays
close to $\alpha$, as Proposition~\ref{prop:conformal} implies, and Risk
rises from 3.6\% to 26.1\% as $\alpha$ grows.
Figure~\ref{fig:conformal_ecr_far} plots the corresponding curves for
the main configuration.

\begin{table}[H]
\centering
\small
\caption{Accept/abstain confusion matrices across $\alpha$ for generator
Olmo-3-32B-Think with the Gemma critic (Spider test split, $n{=}636$).
Positive = accepted.}
\label{tab:olmo_alpha}
\setlength{\tabcolsep}{5pt}
\begin{tabular}{rrrrrccc}
\toprule
$\alpha$ & TP & FP & FN & TN & CondAcc & Recall $=1-$FAR & Risk \\
\midrule
0.1 & 54 & 2 & 416 & 164 & 96.4\% & 11.5\% & 3.6\% \\
0.2 & 90 & 4 & 380 & 162 & 95.7\% & 19.1\% & 4.3\% \\
0.3 & 136 & 6 & 334 & 160 & 95.8\% & 28.9\% & 4.2\% \\
0.4 & 185 & 9 & 285 & 157 & 95.4\% & 39.4\% & 4.6\% \\
0.5 & 239 & 18 & 231 & 148 & 93.0\% & 50.9\% & 7.0\% \\
0.6 & 293 & 27 & 177 & 139 & 91.6\% & 62.3\% & 8.4\% \\
0.7 & 347 & 43 & 123 & 123 & 89.0\% & 73.8\% & 11.0\% \\
0.8 & 395 & 68 & 75 & 98 & 85.3\% & 84.0\% & 14.7\% \\
0.9 & 435 & 99 & 35 & 67 & 81.5\% & 92.6\% & 18.5\% \\
1.0 & 470 & 166 & 0 & 0 & 73.9\% & 100.0\% & 26.1\% \\
\bottomrule
\end{tabular}
\end{table}

\subsection{Partial Critic Configurations}
\label{appendix:partial_critic}

In a \emph{partial} configuration, the generator supplies its own
validation verdict and confidence, and the other model supplies only
the alignment and similarity scores. Table~\ref{tab:partial} compares
these with self-critique and with the full Qwen2.5-72B critic. AUARC
stays between 0.905 and 0.915 across all four.

\begin{table}[H]
\centering
\small
\caption{Degrees of generator--critic separation (generator
GPT-OSS-120B, Spider, $\alpha{=}0.1$).}
\label{tab:partial}
\setlength{\tabcolsep}{4pt}
\begin{tabular}{lcccc}
\toprule
 & Self-critique & Gemma partial & Qwen partial & Qwen critic \\
\midrule
AUROC & 0.781 & 0.782 & 0.821 & 0.826 \\
AUARC & 0.913 & 0.905 & 0.913 & 0.915 \\
ECE   & 0.149 & 0.129 & 0.154 & 0.166 \\
TAR   & 0.993 & 1.000 & 0.993 & 0.993 \\
FAR   & 0.904 & 0.896 & 0.902 & 0.896 \\
\bottomrule
\end{tabular}
\end{table}

\subsection{Seed Variability}
\label{appendix:seeds}

For each configuration, we resample the stratified
train/calibration/test split with seeds 0--9. We refit the forest with
hyperparameters fixed to Table~\ref{tab:rf_params} and recompute the
calibration scores.

\begin{table}[H]
\centering
\small
\caption{Mean $\pm$ standard deviation over 10 resampled splits
($\alpha{=}0.1$). Critic rows use generator GPT-OSS-120B; generator rows
use the Gemma critic and the generation runs of the four-backbone study. The low ECE for generator Qwen2.5-72B reflects its
selected forest (Table~\ref{tab:rf_params}).}
\label{tab:seeds}
\setlength{\tabcolsep}{4pt}
\resizebox{\linewidth}{!}{%
\begin{tabular}{lccccc}
\toprule
Configuration & AUROC & AUARC & ECE & TAR & FAR \\
\midrule
Critic: Gemma-4-31B & $0.772 \pm 0.020$ & $0.900 \pm 0.010$ & $0.151 \pm 0.007$ & $0.988 \pm 0.012$ & $0.907 \pm 0.016$ \\
Critic: Qwen3.8-27B & $0.811 \pm 0.019$ & $0.914 \pm 0.008$ & $0.136 \pm 0.009$ & $0.993 \pm 0.006$ & $0.902 \pm 0.012$ \\
Critic: DeepSeek-V4.1-Flash & $0.793 \pm 0.021$ & $0.907 \pm 0.009$ & $0.145 \pm 0.009$ & $0.988 \pm 0.006$ & $0.905 \pm 0.016$ \\
Critic: Nemotron-3-Super & $0.766 \pm 0.023$ & $0.898 \pm 0.011$ & $0.125 \pm 0.011$ & $0.984 \pm 0.011$ & $0.905 \pm 0.018$ \\
Generator: GPT-OSS-120B & $0.772 \pm 0.023$ & $0.909 \pm 0.010$ & $0.133 \pm 0.011$ & $0.992 \pm 0.007$ & $0.899 \pm 0.015$ \\
Generator: Olmo-3-32B & $0.790 \pm 0.016$ & $0.885 \pm 0.008$ & $0.128 \pm 0.006$ & $0.990 \pm 0.006$ & $0.889 \pm 0.020$ \\
Generator: Qwen2.5-72B & $0.785 \pm 0.015$ & $0.911 \pm 0.007$ & $0.032 \pm 0.009$ & $0.983 \pm 0.012$ & $0.895 \pm 0.016$ \\
Generator: Llama-49B & $0.806 \pm 0.012$ & $0.898 \pm 0.007$ & $0.131 \pm 0.007$ & $0.990 \pm 0.008$ & $0.896 \pm 0.021$ \\
\bottomrule
\end{tabular}}
\end{table}

\subsection{Reliability of Individual LLM-Derived Scores}
\label{appendix:score_reliability}

Table~\ref{tab:score_auroc} reports the standalone AUROC of each
LLM-derived score, before fusion. The first four columns are generator
backbones with the Gemma critic (generation runs of the four-backbone
study, so $c_{\mathrm{gen}}$ for GPT-OSS differs from the last four columns). The last four use generator
GPT-OSS-120B and name the critic. In every configuration,
$c_{\mathrm{gen}}$ comes from the generator and the other four scores
come from the critic. All 40 values exceed 0.5 (range 0.525--0.771).

\begin{table}[H]
\centering
\small
\caption{Standalone AUROC of LLM-derived scores on the Spider test split.}
\label{tab:score_auroc}
\resizebox{\linewidth}{!}{%
\begin{tabular}{lcccccccc}
\toprule
 & \multicolumn{4}{c}{Generator (critic Gemma-4-31B)} & \multicolumn{4}{c}{Critic (generator GPT-OSS-120B)} \\
\cmidrule(lr){2-5}\cmidrule(lr){6-9}
Score & GPT-OSS & Olmo & Qwen2.5 & Llama & Gemma & Qwen3.8 & DeepSeek & Nemotron \\
\midrule
$c_{\mathrm{gen}}$ (generation confidence) & 0.739 & 0.697 & 0.658 & 0.704 & 0.715 & 0.715 & 0.715 & 0.715 \\
$v_{\mathrm{sql}}$ (evaluation verdict) & 0.537 & 0.607 & 0.576 & 0.605 & 0.563 & 0.567 & 0.588 & 0.596 \\
$c_{\mathrm{val}}$ (evaluation confidence) & 0.567 & 0.569 & 0.554 & 0.525 & 0.580 & 0.771 & 0.700 & 0.572 \\
$a$ (alignment) & 0.537 & 0.608 & 0.579 & 0.606 & 0.564 & 0.753 & 0.651 & 0.611 \\
$s_{\mathrm{sim}}$ (question similarity) & 0.652 & 0.721 & 0.665 & 0.704 & 0.651 & 0.737 & 0.710 & 0.595 \\
\bottomrule
\end{tabular}}
\end{table}

\subsection{Error Analysis}
\label{appendix:error_analysis}

The feature distributions (Appendix~\ref{appendix:feature_distribution})
show where individual signals fail. Logit features overlap heavily
between correct and incorrect queries, because a wrong query is
usually still fluent token by token. The error is semantic, not a
decoding hesitation. The critic's verdict and alignment score peak at
their ``looks correct'' value for most correct queries, and most
incorrect queries share that peak.

We examined the 144 incorrect test queries of the main configuration,
where GPT-OSS-120B generates and Gemma-4-31B critiques. We call an
error \emph{recognized} if the critic's verdict $v_{\mathrm{sql}}$ and its
alignment score $a$ both flag it ($v_{\mathrm{sql}} = 0$ and $a < 50$;
20 queries), and \emph{not recognized} if neither does (124 queries).
The critic therefore misreads most errors. Recognized errors receive a
mean forest posterior of $p = 0.34$, and at most one is accepted at
any operating level up to $\alpha{=}0.5$. Unrecognized errors receive a
higher mean posterior ($p = 0.47$), yet only 3 of 124 are accepted at
$\alpha{=}0.1$, because the generator-side features still separate them
from most correct queries. As $\alpha$ increases to 0.5, up to 14 of
124 (11\%) are accepted. Errors that the critic misreads are thus
the main source of accepted errors at larger $\alpha$.

\subsection{Zero-Shot Transfer to BIRD}
\label{zero_shot_results}

\begin{table}[H]
\centering
\small
\caption{Zero-shot transfer to the BIRD dev set (1,534 queries, generator
GPT-OSS-120B, critic Gemma-4-31B, Spider-trained forest and
calibration, no retraining or recalibration). Precision, recall, and
F1 are for error detection ($\hat{c} < 0.9$). AUARC ranges from 0.343
(random) to 0.709 (perfect ranking) at this base accuracy.}
\label{tab:bird}
\begin{tabular}{lc}
\toprule
\textbf{Metric} & \textbf{Value} \\
\midrule
Base accuracy        & 0.342 \\
ECE                  & 0.091 \\
ACE                  & 0.090 \\
Brier                & 0.202 \\
AUROC                & 0.724 \\
AUARC                & 0.515 \\
Error-detection precision & 0.662 \\
Error-detection recall    & 0.998 \\
Error-detection F1        & 0.796 \\
CondAcc ($\alpha{=}0.1$)  & 0.833 \\
Risk at Spider-certified $\hat{\lambda}$ ($\varepsilon{=}0.15$) & 0.301 \\
\bottomrule
\end{tabular}
\end{table}

\label{appendix:bird_ece}
\textbf{Why ECE is lower on BIRD than on Spider.} The lower ECE does not
mean better calibration out of domain. It is a base-rate effect
(Table~\ref{tab:bird_ece}). On Spider, base accuracy is high, so even
low-confidence predictions are often correct, and empirical accuracy
exceeds confidence in every bin. On BIRD, most predictions have low
confidence, and in that region accuracy is close to mean confidence,
so it contributes little to ECE. Conformal
calibration explains neither result. It concerns the threshold, not
the binned agreement that ECE measures, and its guarantee does not
extend to BIRD, whose queries are not exchangeable with the Spider
calibration set.

\begin{table}[H]
\centering
\small
\caption{Base-rate comparison behind the ECE difference (generator
GPT-OSS-120B, critic Gemma-4-31B).}
\label{tab:bird_ece}
\begin{tabular}{lcc}
\toprule
 & Spider (in-domain) & BIRD (zero-shot) \\
\midrule
Error rate without abstention & 21.5\% & 65.8\% \\
Share of predictions with $p \leq 0.5$ & 29\% & 72\% \\
Accuracy vs.\ mean confidence there & 53\% vs.\ 36\% & 24\% vs.\ 35\% \\
ECE                           & 0.133 & 0.091 \\
\bottomrule
\end{tabular}
\end{table}

\subsection{Latency Breakdown}
\label{latency_breakdown}

\begin{table}[H]
\centering
\small
\caption{Per-query wall-clock latency (seconds) of the main configuration.
The generator, critic and local stages were timed one query at a time on
40 test queries. Stage~3 can run in parallel with Stage~2.}
\label{tab:latency}
\begin{tabularx}{\textwidth}{l l Y}
\toprule
\textbf{Stage} & \textbf{Operation} & \textbf{Mean time (s)} \\
\midrule
1 & SQL generation (GPT-OSS-120B) + execution & 1.9 - 2.6 \\
2 & Validation + alignment (Gemma-4-31B critic) & 0.5 - 1.7 \\
3--4 & Features, forest, conformal score (local) & 0.03 \\
\midrule
\textbf{Total} & per query & \textbf{2.43 - 4.33} \\
\bottomrule
\end{tabularx}
\end{table}

The generator dominates the cost, because of the query execution. Both LLM calls ran through a shared
inference gateway with reasoning enabled. Feature extraction, forest
training and inference, and conformal calibration ran on a MacBook M4
Pro without a GPU.

\subsection{Published Results of Existing Methods}
\label{appendix:compare_with_existing_methods}

Table~\ref{tab:comparison_existing} lists results reported in the
original papers on Spider. These use different backbones, splits, and
evaluation protocols, so they are \emph{not} directly comparable with
ours. Three of the four report AUROC but not AUARC, so AUROC is the
closest common metric. Table~\ref{tab:controlled} gives the controlled
comparison.

\begin{table}[H]
  \caption{Reported results of existing confidence estimation methods on
  Spider, each on its own backbone and protocol.}
  \label{tab:comparison_existing}
  \centering
  \small
  \begin{tabularx}{\textwidth}{L{4.8cm} L{2.8cm} L{2.0cm} Y}
    \toprule
    \textbf{Method} & \textbf{Backbone} & \textbf{Approach} & \textbf{Reported results} \\
    \midrule
    SAC-Product~\citep{entezari2025confidence} & GPT-4o-mini & Logit-based & AUROC 79.9, ECE 0.14 \\
    Geometric Aggregation~\citep{xiao2025consistency} & CodeLlama-34B & Black-box & AUROC 76.0, AUARC 35.0 \\
    Multivariate Platt Scaling~\citep{liu2025calibrating} & Llama-3.1-8B & Logit + sampling & AUROC 74.8, Brier 0.163, ECE 0.024, ACE 0.026 \\
    Pooled token-level~\citep{ramachandran2024text} & Codestral-22B & Logit-based & AUROC 78.8 \\
    \midrule
    Ours & Olmo-3-32B-Think + Gemma-4-31B critic & Combined & AUARC 89.5, AUROC 81.6, ECE 0.126, ACE 0.125, Brier 0.158 \\
    \bottomrule
  \end{tabularx}
\end{table}

\section{Applicability Beyond Text-to-SQL}
\label{appendix:generalization}

Parts of the pipeline do not depend on SQL. These are the
generator--critic separation, the critic's validation step, the
token-level features, the forest
fusion, and the conformal calibration. They apply to any task that
produces an output with token probabilities and has labeled data on
which correctness can be computed. Stage~2's alignment step and the
structural features are SQL-specific. The alignment step needs a concrete result, a way to generate
the question the result answers, and a check of that result. Code
generation (with unit tests as the execution analogue) and
mathematical reasoning (with a solver) have such analogues. Open-domain
QA does not, and Stage~2 would need a replacement such as an
attribution check of whether retrieved evidence supports the answer.
We have not evaluated these extensions.

\end{document}